\documentclass[lettersize,journal]{IEEEtran}
\usepackage{amsmath,amsfonts}
\usepackage{array}
\usepackage[caption=false,font=normalsize,labelfont=sf,textfont=sf]{subfig}
\usepackage{textcomp}
\usepackage{stfloats}
\usepackage{url}
\usepackage{verbatim}
\usepackage{cite}
\usepackage[ruled,linesnumbered]{algorithm2e}
\usepackage{booktabs}
\usepackage{enumitem}
\usepackage{multirow}
\usepackage{tabularx}
\usepackage{graphicx}
\usepackage{stackengine}

\usepackage{hyperref}

\begin{document}

\title{A Low-Computational Video Synopsis Framework with a Standard Dataset}

\author{\href{https://orcid.org/0009-0001-0599-4118}{\includegraphics[scale=0.06]
		{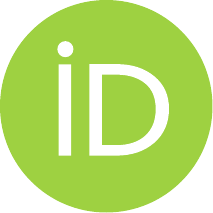}\hspace{1mm}R. Malekpour},
	\href{https://orcid.org/0000-0003-2237-1374}{\includegraphics[scale=0.06]
		{orcid.pdf}\hspace{1mm}M. M. Morsali}, 
	\href{https://orcid.org/0000-0002-9852-5088}{\includegraphics[scale=0.06]
		{orcid.pdf}\hspace{1mm}H. Mohammadzade*\thanks{*Corresponding author, email: hoda@sharif.edu}}\\	
	Department of Electrical Engineering, Sharif University of Technology}

\markboth{Arxiv preprint, September 2024}
{Malekpour \MakeLowercase{\textit{et al.}}: A Comprehensive Framework for Video Synopsis with Enhanced Algorithms and a Standard Dataset}

\maketitle

\begin{abstract}
Video synopsis is an efficient method for condensing surveillance videos. This technique begins with the detection and tracking of objects, followed by the creation of object tubes. These tubes consist of sequences, each containing chronologically ordered bounding boxes of a unique object. To generate a condensed video, the first step involves rearranging the object tubes to maximize the number of non-overlapping objects in each frame. Then, these tubes are stitched to a background image extracted from the source video. The lack of a standard dataset for the video synopsis task hinders the comparison of different video synopsis models. This paper addresses this issue by introducing a standard dataset, called SynoClip, designed specifically for the video synopsis task. SynoClip includes all the necessary features needed to evaluate various models directly and effectively. Additionally, this work introduces a video synopsis model, called FGS, with low computational cost. The model includes an empty-frame object detector to identify frames empty of any objects, facilitating efficient utilization of the deep object detector. Moreover, a tube grouping algorithm is proposed to maintain relationships among tubes in the synthesized video. This is followed by a greedy tube rearrangement algorithm, which efficiently determines the start time of each tube. Finally, the proposed model is evaluated using the proposed dataset. The source code, fine-tuned object detection model, and tutorials are available at \url{https://github.com/Ramtin-ma/VideoSynopsis-FGS}.
\end{abstract}


\section{Introduction}
\IEEEPARstart{T}{he} use of surveillance cameras recording continuously is increasing, aiding in public safety and security. However, their constant recording leads to high-volume redundant video content, complicating storage and review processes. To address this issue, various methods have been introduced to produce condensed videos that retain crucial information. These condensed videos not only reduce storage requirements but also simplify the video review process. There are mainly two approaches to produce a condensed video: video synopsis and video summarization. 

\begin{figure*}[!t]
	\centering
	\includegraphics[width=\linewidth]{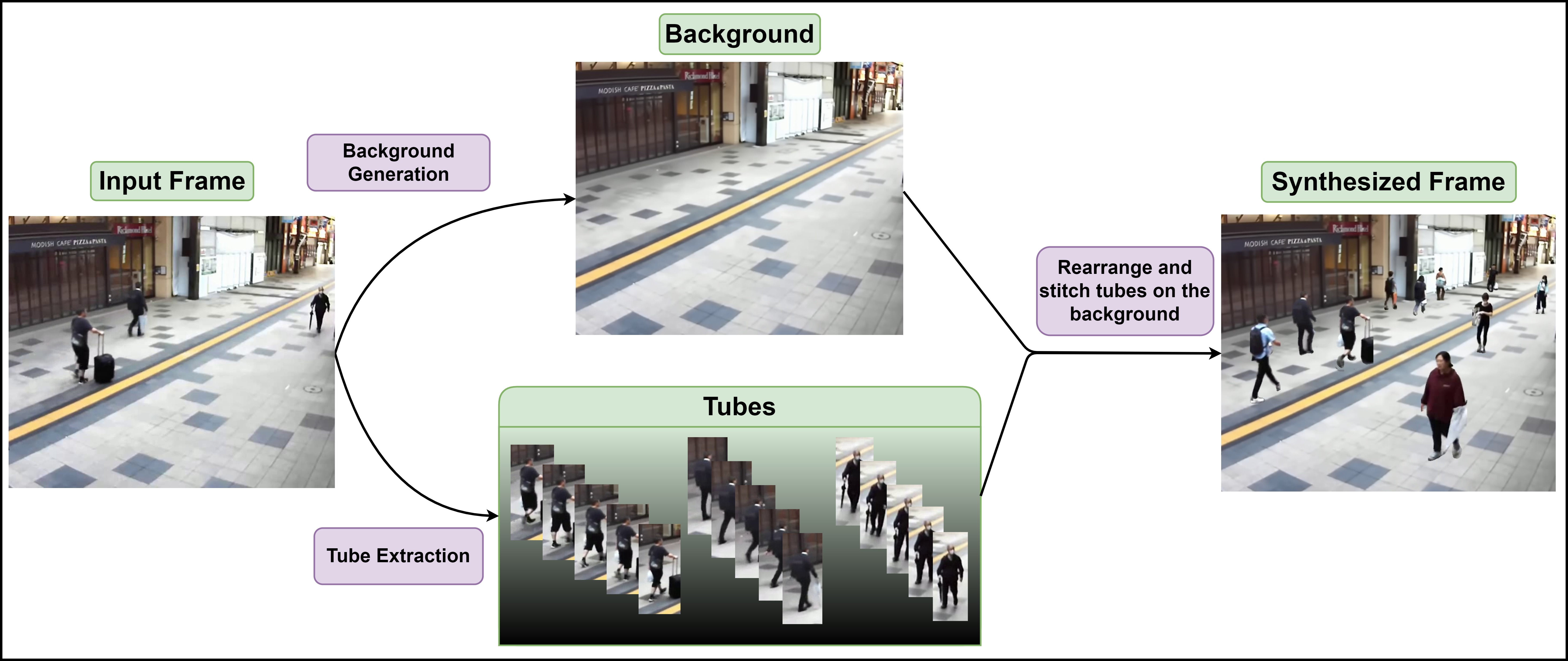}
	\caption{Condensed video production using the video synopsis approach.}
	\label{Synopsis Framework}
\end{figure*}

In video summarization, keyframes are extracted from the video to create a condensed video. While this approach can be applied to various types of videos, it results in some information loss, as it always discards some frames\cite{VS1, VS2, VS3, VS4, VS5, VS6, VS7}. Video synopsis, on the other hand, is specifically designed for stationary cameras with a static background. It assumes that the video contains moving objects against a consistent background. In this approach, as illustrated in Figure \ref{Synopsis Framework}, the background is extracted, and all moving objects are detected and tracked to create object tubes. These tubes are then placed in the background to produce a compressed video that retains all information from the original video. While this approach is limited to videos with a static background, it has the capability to preserve all the events and information in the original video.

Video synopsis confronts a critical challenge in maintaining the information associated with the relationship of tubes. In videos, it is common to see individuals chatting or walking together. Preserving the relationship between those tubes is crucial for accurately representing events, as their separation in the condensed video leads to the loss of important contextual information. Therefore, detecting the connections between tubes and preserving their time interval in the condensed video is essential for retaining crucial information. While early research in video synopsis overlooked this issue, recent studies\cite{Ruan, Yang, G1, Zhu, Fu} have proposed solutions to address it. Ruan \textit{et al.} \cite{Ruan} modeled tube relationship with a dynamic graph, where nodes represent object masks of tubes and edges represent the relationship between them. Based on scene geometry, spatial distance, and the height of tubes, Yang \textit{et al.} \cite{Yang} proposed a relationship function, detecting relationships by considering the duration of the interaction.

Tube rearrangement is a crucial step in video synopsis, aiming to maximize compression while minimizing information loss resulting from tube collisions. Tube rearrangement algorithms can be categorized into online, offline, and dynamic algorithms.

Offline algorithms begin by extracting all the tubes from the input video. Subsequently, adopting a global optimization strategy, they determine the start time of each tube \cite{Pritch, Ghatak, swarm, Li, Nie, Nie2, Chen, G1}. Pritch \textit{et al.} \cite{Pritch} was one of the first researchers to devise a global energy function to compute the tube's start time. Ghatak \textit{et al.} \cite{Ghatak} propose a hybrid algorithm, named SA-JAYA, for energy minimization to achieve an optimal solution with a faster convergence rate. Additionally, they employ the Analytic Hierarchy Process (AHP) for weight assignment, enhancing reliability over heuristic methods. Moussa \textit{et al.} \cite{swarm} employed a particle swarm optimization algorithm to rearrange tubes. Other researchers have investigated alternative approaches, including modifying the size, speed, and position of tubes to reduce collisions. For instance, Li \textit{et al.} \cite{Li} initially positioned the tubes and then scaled down tubes that collided. Nie \textit{et al.} \cite{Nie} introduced an optimization problem that dynamically adjusts the tube's start time, speed, and scale to minimize collisions. Additionally, Nie proposed a spatiotemporal optimization approach allowing tubes to be repositioned in both space and time \cite{Nie2}.

Online algorithms process tubes in sets during the tube extraction stage \cite{Zhu, Huang, Jin, Feng, Fu, He, Ra}. Consequently, online algorithms encounter a stepwise optimization problem, often tackled using greedy algorithms to determine a tube's start time. Zhu \textit{et al.} \cite{Zhu} introduced an online tube-filling algorithm inspired by the Tetris game. This algorithm treats each tube as a tetromino, striving to occupy the 3D space of the video. Huang \textit{et al.} \cite{Huang} formulated the tube rearrangement problem as a maximum a posteriori probability (MAP) estimation problem, enabling the rearrangement of tubes without complete trajectories. Jin \textit{et al.} \cite{Jin} proposed a projection strategy using an associated projection matrix that holds the latest information of the video space. Their method iteratively rearranges tubes and updates the projection matrix, eliminating the need for direct tube comparison.

In online methods, once tubes are positioned, their locations remain fixed regardless of subsequent tubes, making offline methods generally more effective in compression due to their access to information from all tubes. However, offline methods cannot directly process video streams since they require video segments as input. Conversely, online algorithms are capable of directly processing video streams.

Recently, researchers have explored dynamic algorithms to leverage the advantages of both online and offline methods. Similar to online approaches, these algorithms process tubes in sets. However, the introduction of new tubes might trigger the rearrangement of previously positioned tubes. Ruan \textit{et al.} \cite{Ruan} developed a dynamic graph coloring algorithm to rearrange tubes. Adding a new tube to this graph could potentially alter the start times of previous tubes. Yang \textit{et al.} \cite{Yang} employed the octree algorithm for dynamic tube placement. This algorithm defines R1 and R2 spaces, where tubes in R1 maintain their optimal positions while tubes in R2 can be adjusted based on upcoming tubes.

One major challenge in assessing video synopsis models is the absence of a standard dataset. As a result, directly comparing the performance of various models based on measurements from different studies is not feasible. Many researchers have relied on private datasets, which are not publicly available, for model evaluation. Others have used surveillance camera videos or tracking datasets such as PETS 2009 \cite{PETS}, CAVIAR \cite{CAVIAR}, Hall monitor \cite{Hall}, Day-time \cite{Daytime}, F-building \cite{F}, KTH \cite{KTH}, WEIZMAN \cite{WEIZMAN}, VIRAT \cite{VIRAT}, and Sherbrooke Street \cite{Sherbrooke}. However, none of the aforementioned datasets have all the features required for the video synopsis task, which leads to no agreement for the video synopsis dataset. This issue has been highlighted in several review articles\cite{D1, D2, D3}. Additionally, Ingle \textit{et al.} \cite{D4} conducted an experiment to evaluate the performance of different models under identical conditions, a time-consuming process that could be avoided by introducing a standard video synopsis dataset.

A video synopsis dataset must include 4 features.

\begin{enumerate}[leftmargin=*]
	\item \textbf{Videos should not be crowded:} In crowded videos, most frames are already occupied by objects, leaving little space to add more. Consequently, the video synopsis approach may not achieve significant compression in such cases.
	\item \textbf{The length of the videos should be long:} Short videos may not provide an adequate number of tubes for proper evaluation of synopsis models.
	\item \textbf{The camera must be stationary:} The video synopsis task assumes a static background. If the camera is in motion, this condition will not be met.	
	\item \textbf{Videos must contain tube annotations:} To evaluate tube rearrangement algorithms, it is essential to use identical tubes. Different tube extraction algorithms may yield different tubes for the same video without annotations, hindering direct comparison of tube rearrangement algorithms.
\end{enumerate}

The key contributions of this paper include:

\begin{itemize}[leftmargin=*]
	\item \textbf{A Video Synopsis Dataset:} This paper introduces a dataset, called SynoClip, specifically designed for video synopsis, encompassing all the aforementioned features.
	\item \textbf{A Fast Tube Rearrangement Algorithm:} This paper presents an efficient tube rearrangement algorithm, inspired by \cite{G1}, called Fast Greedy Synopsis (FGS), which surpasses the original algorithm in terms of computational efficiency and compression performance.
	\item \textbf{An Empty-Frame Object Detector:} This work introduces an empty-frame object detector to reduce computational costs, preventing the utilization of deep models for frames empty of any desired objects.
	\item \textbf{A Segmentation Algorithm:} This paper presents a segmentation algorithm designed to remove background pixels from each object's bounding box, thereby enhancing the quality of the output videos.
\end{itemize}

\section{Proposed Method}
\begin{figure*}[!t]
	\centering
	\includegraphics[width=\linewidth]{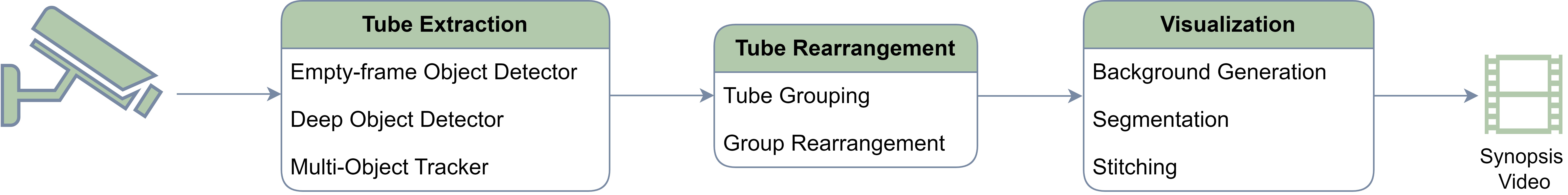}
	\caption{The proposed video synopsis system.}
	\label{System}
\end{figure*}

Figure \ref{System} illustrates an overview of the proposed video synopsis system, consisting of three main modules: tube extraction, tube rearrangement, and visualization. Each module incorporates different units.

The tube extraction module comprises three units: a deep object detector, an empty-frame object detector, and a multi-object tracker. Typically, the deep object detector identifies objects in a video frame, while the multi-object tracker assigns a unique ID to each detected object and maintains ID consistency across frames. If the deep object detector reports no objects in a frame, the system seamlessly switches to the empty-frame object detector, a computationally lightweight alternative. Once the empty-frame object detector detects a desired object, the system reverts to using the deep object detector. This dynamic switching technique optimizes computational efficiency. The resulting tubes represent the spatiotemporal trajectories of each object in the video.

After the tube extraction, the tube rearrangement module integrates the tubes into the synopsis video. Often, there are some connections between the objects in a video; for instance, people walking together or engaging in conversation. The tube grouping unit organizes the tubes into groups, placing related tubes within the same group to preserve their connection. This ensures a simultaneous display of related tubes in the synopsis video. The optimization unit determines the timing for each tube to appear in the synopsis video, aiming to minimize collisions and maximize video compression.

In the visualization module, the synthesis of the synopsis video takes place. The background generation unit creates a background by utilizing sample frames collected randomly from the input video. The segmentation unit processes the object images from the tubes, eliminating background pixels to enhance the output video quality. The stitching unit strategically positions the tube images on the background at the designated times determined by the tube rearrangement module.

The subsequent subsections elaborate on the individual units within the system modules.

\subsection{Tube Extraction}

\subsubsection{Deep Object Detector}
The ideal deep object detector for the video synopsis should demonstrate fast processing speed, high detection accuracy, and minimal computational resource consumption. These criteria narrow down the options to YOLO-like detectors. Wang \textit{et al.} \cite{yolo7} provides a comprehensive comparison of YOLO-like state-of-the-art object detectors. Furthermore, another study \cite{yolo8} compares the latest version of the YOLO object detector with its predecessors. Based on these studies, YOLOv8n emerges as a promising choice. To enhance detection accuracy, a preprocessing step involves resizing all input videos to a specific dimension equal to the geometric mean of the dataset videos. Moreover, a subset of video frames is carefully selected for fine-tuning the object detector.

Given the high frame rate of contemporary videos, the displacement of objects across consecutive frames tends to be minimal. Therefore, this paper employs a detection-with-stride technique to reduce computational workload. After detecting objects in a frame, the subsequent two frames are excluded from processing. Instead, the object locations in these frames are estimated by interpolating bounding boxes from the first and fourth frames.

\subsubsection{Empty-frame Object Detector}
In cases where video frames lack objects of interest, relying on a computationally expensive deep object detector appears inefficient. The system defaults to using the deep object detector. However, upon encountering an empty frame, the system temporarily disables the deep object detector and transitions to the empty-frame object detector. This specialized detector rapidly analyzes the frame to identify any potential object presence. Once the empty-frame object detector signals the presence of objects in a frame, the system reactivates the deep object detector to process the current frame and subsequent frames.

\begin{figure*}[!t]
	\centering
	\includegraphics[width=\linewidth]{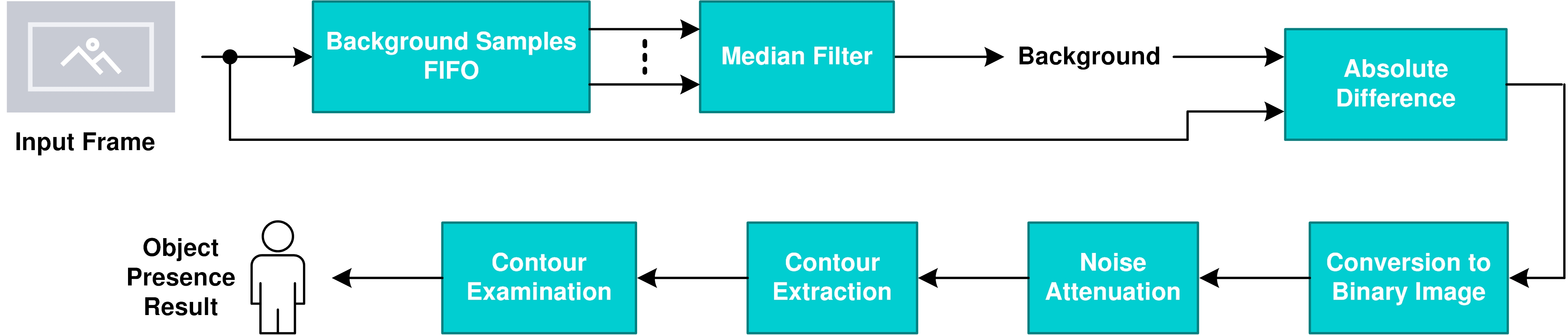}
	\caption{Processing steps of the proposed empty-frame object detector.}
	\label{Empty-frame}
\end{figure*} 

Figure \ref{Empty-frame} illustrates the processing steps of the proposed empty-frame object detector. The process begins with filling the background samples FIFO. This FIFO holds ten samples of empty frames, meaning frames devoid of objects of interest. Initially, the system fills the FIFO with a sample of an empty frame identified by the deep object detector. Subsequently, it periodically adds new samples of empty frames, identified by the empty-frame object detector or deep object detector, to the FIFO. New samples introduced to this FIFO will replace the oldest sample currently residing within it.

The empty-frame object detector periodically applies a pixel-wise median filter on the samples stored in the FIFO to generate a new background. This median filter fills each pixel in the background with the median value of the corresponding pixels in the empty frame samples. This updating process is essential due to potential changes in background features, such as lighting, resulting from the prolonged absence of desired objects in the video. While a higher update rate enhances the background's quality, it also intensifies computational costs.

After generating the background, the system computes the absolute difference between the input frame and the background. The resulting difference image is then converted into a binary image using a suitable threshold. Subsequently, morphological operations are applied to eliminate noise from the binary image. The system then proceeds to detect object contours in the binary image. If the system detects an object contour with an area and aspect ratio matching those of the desired object contour, it considers the input frame as containing objects; otherwise, it regards it as empty.

\subsubsection{Multi-object Tracker}
In a video synopsis system, an ideal multi-object tracker should exhibit rapid response with low computational cost. After reviewing existing literature on this matter, this paper has chosen SFSORT, as proposed in \cite{sfsort}, due to its ability to deliver superior accuracy in real-time with minimal computations. The tracker receives predictions and their corresponding detection scores from the object detector, YOLOv8n, and assigns a unique ID to each valid detection. The IDs of objects present in previous frames remain consistent, ensuring tracking continuity.

\subsection{Tube Rearrangement}
As previously discussed, the time interval of related tubes must be maintained in the synopsis video. 
Additionally, in cases where one object occludes another, a portion of an object appears within the images of another tube. Positioning the tubes of occluding objects at different times will reduce the quality of the synopsis video. Preserving the time interval between these tubes will naturally represent the occlusion of these tubes, as observed in the source video. So, to generate a high-quality synopsis video, the tube grouping algorithm needs to account for both related and occluding tubes.

\subsubsection{Tube Grouping}
\begin{figure*}[!t]
	\centering
	\begin{tabular}{ccc}
		\includegraphics[width=0.3\textwidth]{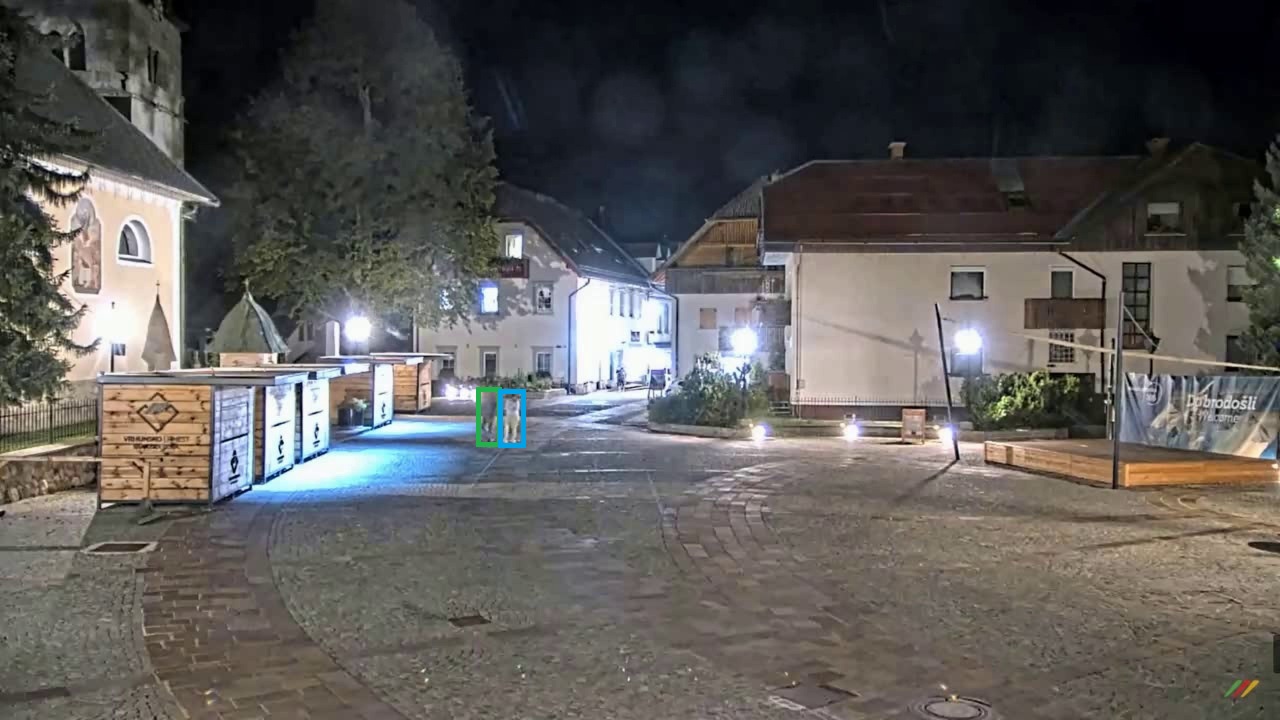} & 
		\includegraphics[width=0.3\textwidth]{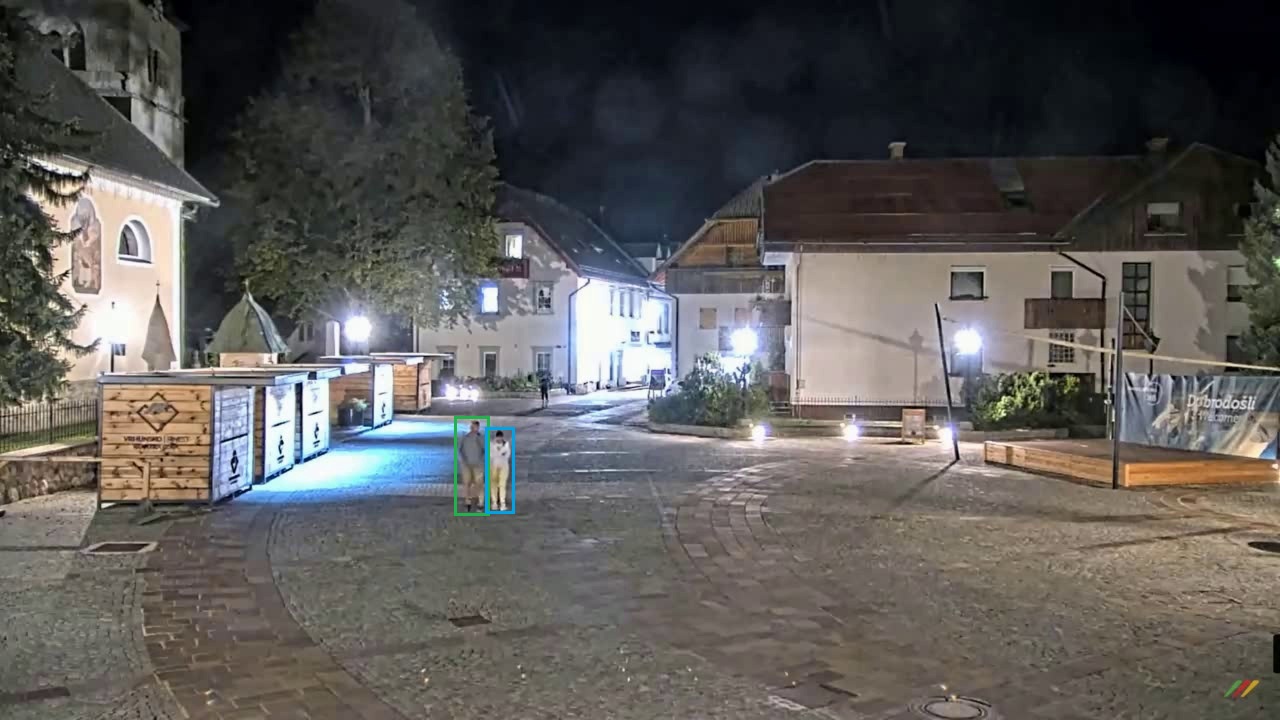} & 
		\includegraphics[width=0.3\textwidth]{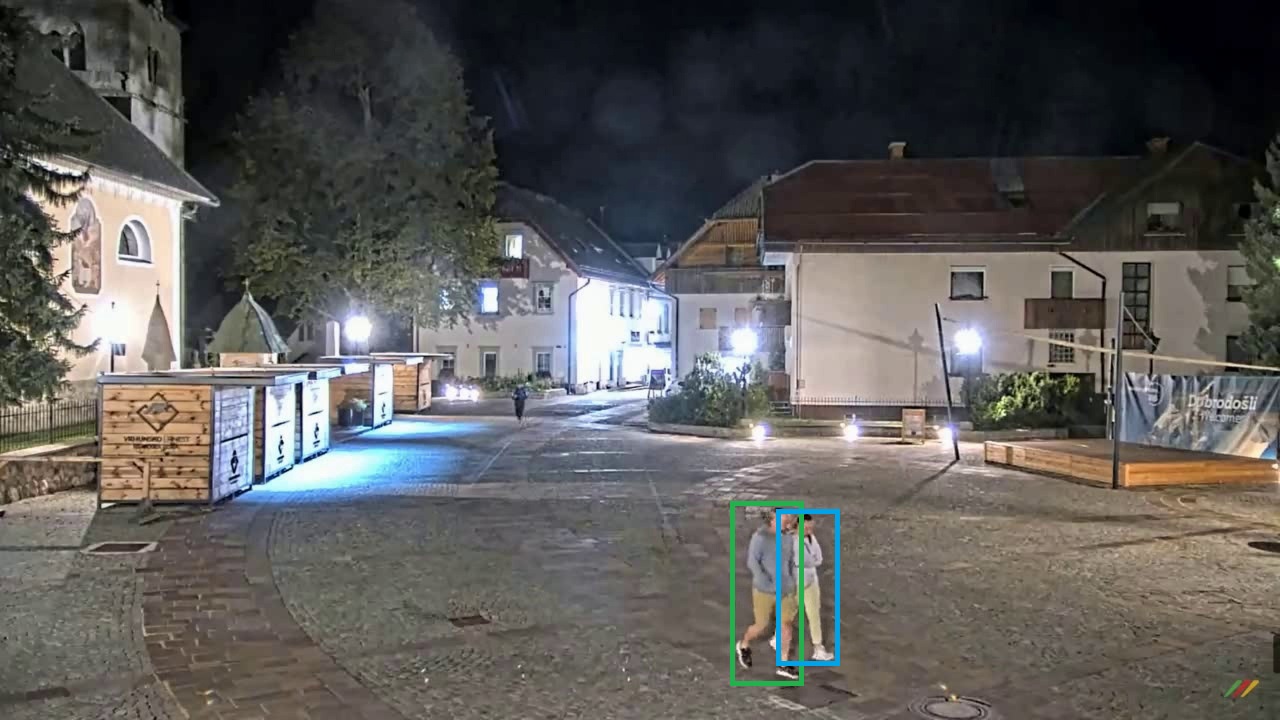} \\
		 & (a) & \\
		\includegraphics[width=0.3\textwidth]{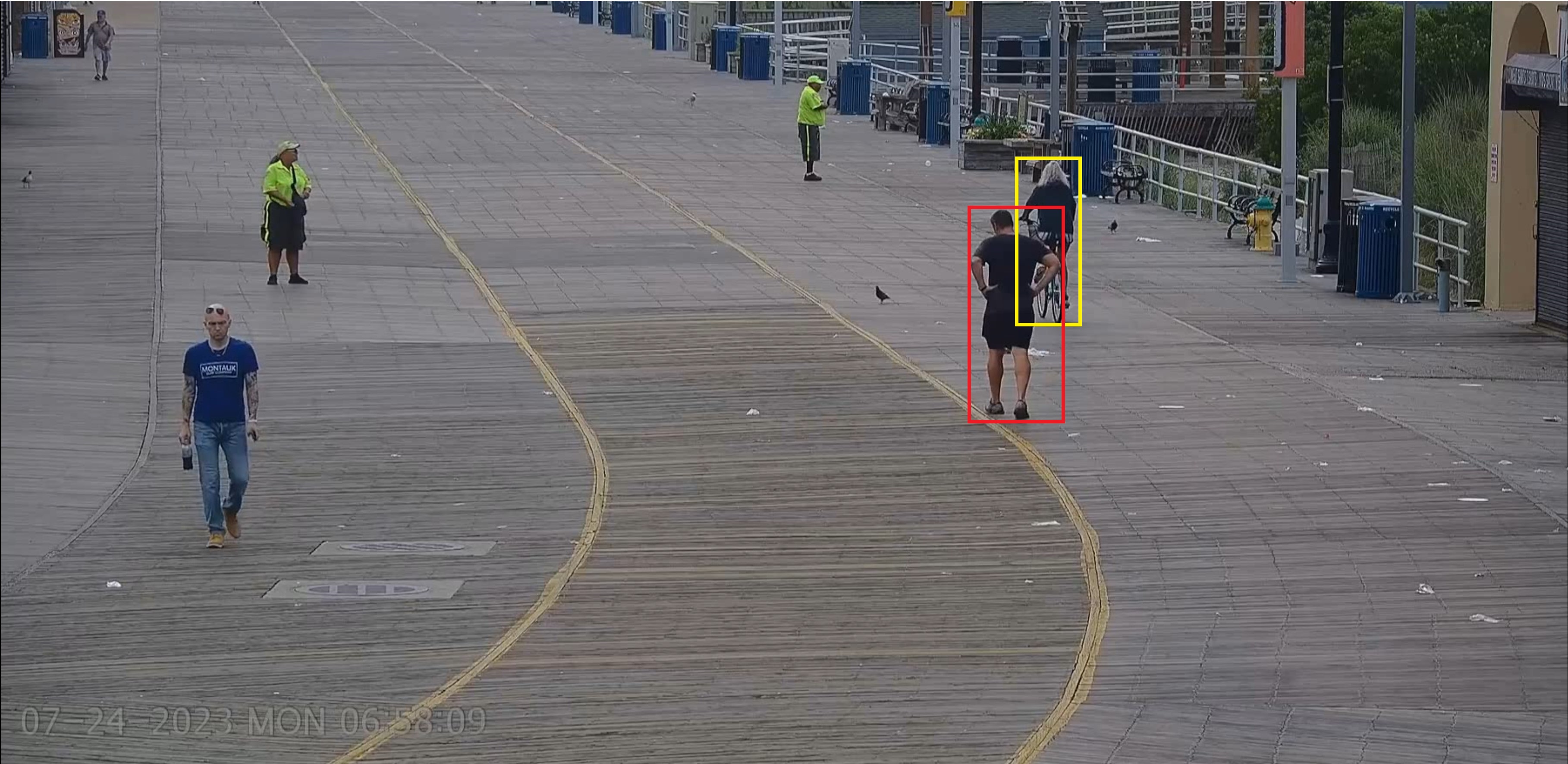} & 
		\includegraphics[width=0.3\textwidth]{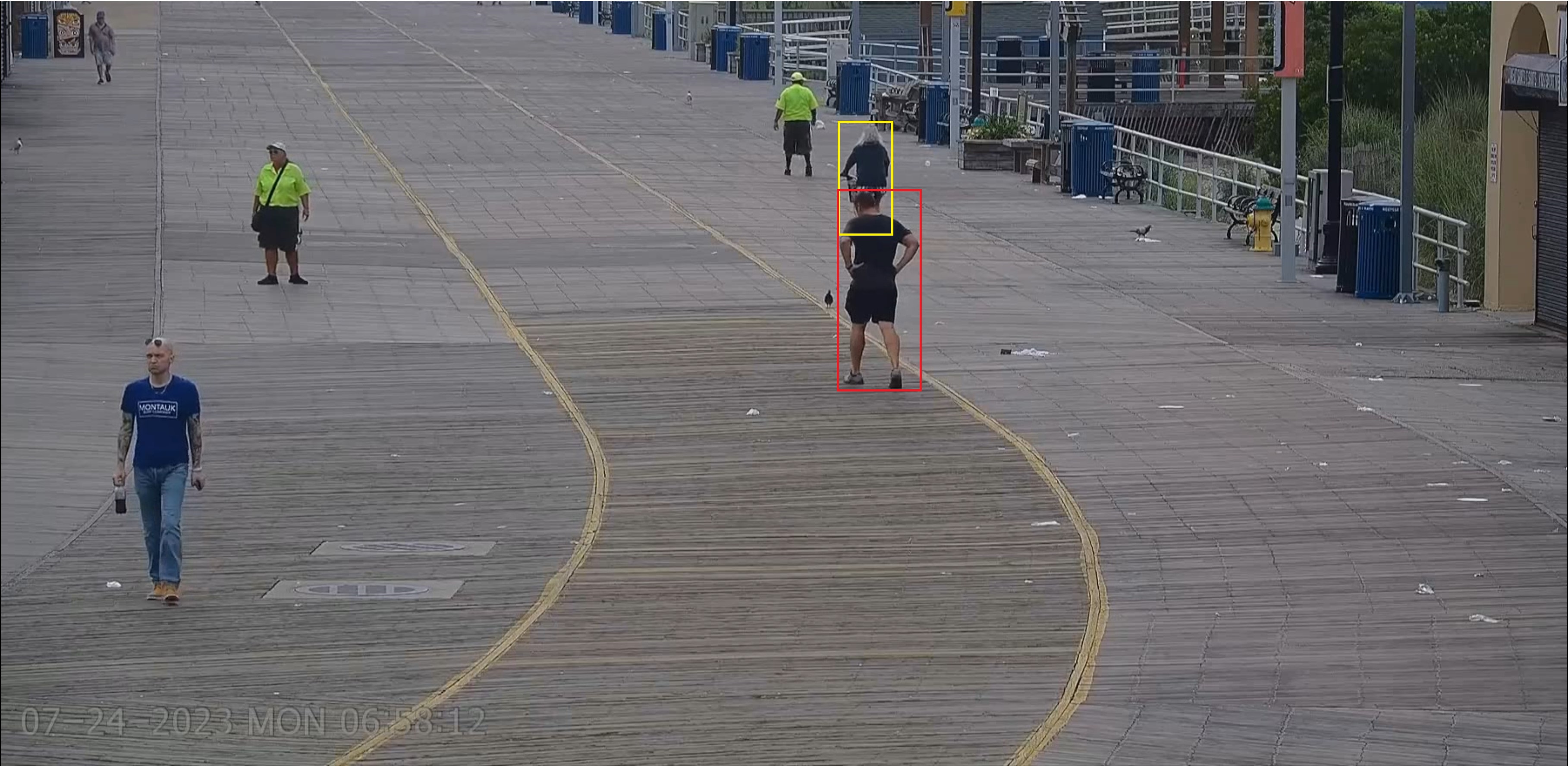} & 
		\includegraphics[width=0.3\textwidth]{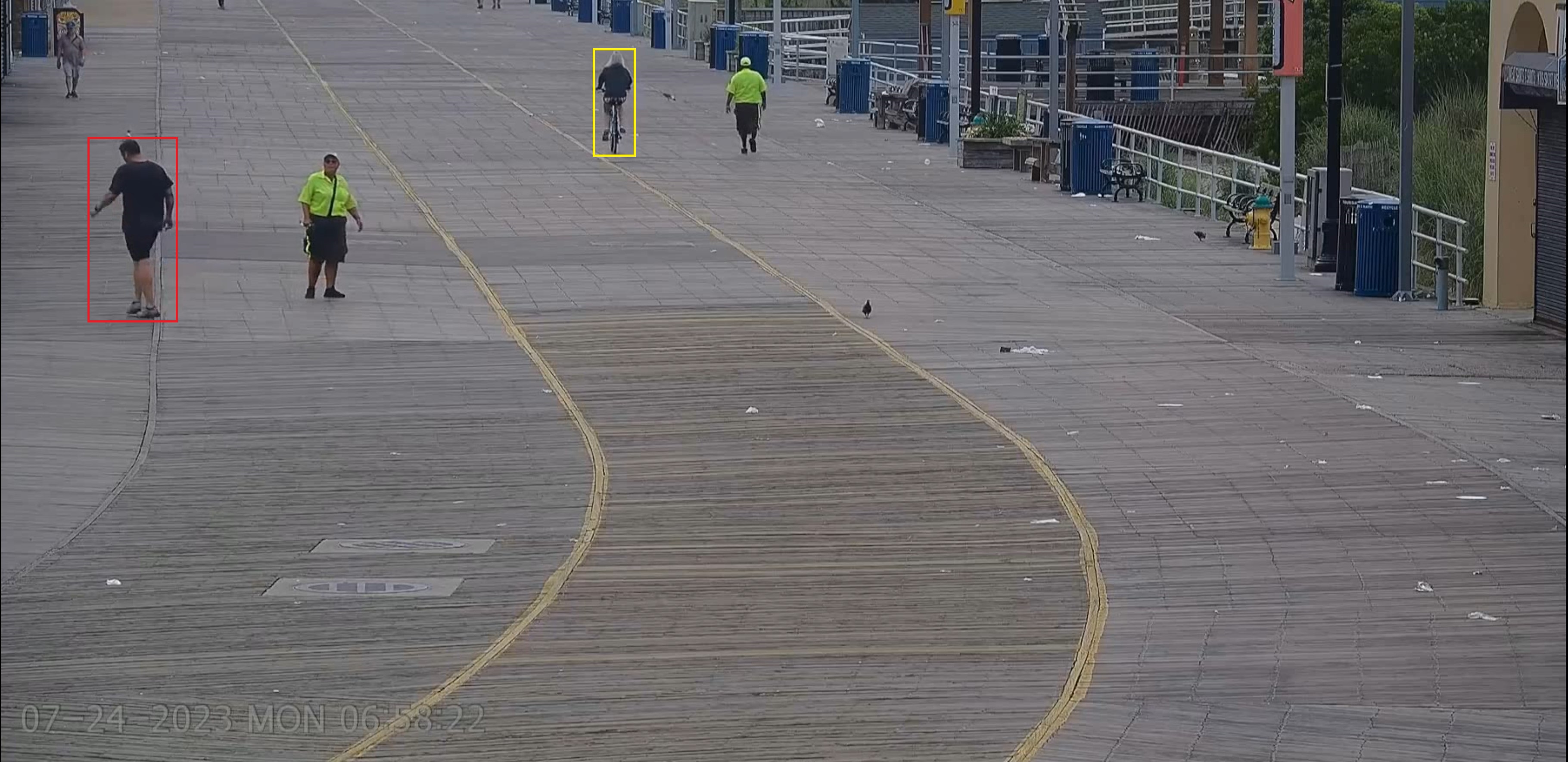} \\
		& (b) & \\
	\end{tabular}
	\caption{Examples illustrating the necessity of tube grouping based on spatial distance and occlusion: (a) Two interacting tubes (green and blue boxes) with low average spatial distance in shared frames. (b) Two tubes (red and yellow boxes) with distinct trajectories but frequent overlaps.}
	\label{Collision:fig}
\end{figure*}

A pair of tubes will be placed in the same group under two conditions: either they exhibit a low average spatial distance throughout the video, or they experience a high number of overlapping occurrences. Figure \ref{Collision:fig} illustrates an example of these two scenarios in which tubes are grouped.

A relatively small average distance between two tubes in their common frames, where both tubes were present, signifies a probable interaction between them. So, Equation \ref{Grouping_distance} defines the average distance of two tubes:

\begin{equation}
	\label{Grouping_distance}
	D(i,j) =  \frac{\sum_{t \in t_i \cap t_j} d(B_i^t,B_j^t)}{Com(T_i,T_j)},
\end{equation}

\noindent where $T$ represents a tube, $B_i^t$ indicates the bounding box of tube $i$ in frame $t$, $d(B_i^t,B_j^t)$ denotes the center-to-center distance, and $Com(T_i,T_j)$ represents the number of common frames between tubes $i$ and $j$.

A low average distance alone is not sufficient to conclude that there is a relationship between two tubes because two nearly non-concurrent tubes may exhibit a low average distance in the few common frames they share. Therefore, the average distance should be weighted using a weight function to increase the distance between non-concurrent tubes, thus preventing non-concurrent tubes from being mistakenly identified as related. Equation \ref{Grouping_weight} illustrates how the weighting occurs for each pair of tubes:

\begin{equation}
\label{Grouping_weight}
W(i,j) = f\left(\frac{Com(T_i,T_j)}{min(T_i,T_j)}\right),
\end{equation}

\noindent where $min(T_i,T_j)$ indicates the minimum number of frames containing either tube $i$ or tube $j$, and $f$ is an arbitrary injective decreasing function in the range of $(0, 1)$. The fraction inside the function $f$ describes the concurrency of tubes $i$ and $j$ by a number ranging from $0$ for non-concurrent tubes to $1$ for concurrent tubes. Equation \ref{f} illustrates an example of the function $f$ employed in this study:

\begin{equation}
	\label{f}
	f(x) = (1 + \frac{1}{1+e^{x/2}})^4.
\end{equation}

Finally, two tubes will be placed in the same group if their weighted average distance, as defined in Equation \ref{Weight_distance}, is less than a specific threshold.

\begin{equation}
	\label{Weight_distance}
	DW(i,j) = D(i,j) \times W(i,j).
\end{equation}

To identify highly overlapping tubes, Equation \ref{Group occlusion} computes the sum of the Intersection over Minimum (IoM) between the bounding boxes of two tubes, serving as a measure of collision between tubes.

\begin{equation}
\label{Group occlusion}
C(i,j) = \sum_{t \in t_i \cap t_j} \frac{I(B_i^t,B_j^t)}{min(B_i^t,B_j^t)}.
\end{equation}

Here, $I(B_i^t, B_j^t)$ denotes the intersection area between the bounding boxes of tubes $i$ and $j$ in frame $t$, while $min(B_i^t,B_j^t)$ represents the area of the smaller bounding box. The normalization achieved through division by $min(B_i^t,B_j^t)$ helps mitigate the impact of box size on collision cost. If the collision frequency, described by the sum of IoM, between two tubes, exceeds a certain threshold, those two tubes will be grouped together.

The grouping method should feature the transitive property: Any pairs of tube groups like (a, b) and (b, c), which share common members, should merge to form an extended group like (a, b, c). Consequently, after detecting all pairs of tubes, the grouping method should merge certain groups.

The proposed grouping method, outlined in Algorithm \ref{alg:grouping}, takes the list of tubes and two threshold values as input. The algorithm first identifies all pairs of tubes that meet either the weighted distance or collision criteria in lines $3$ to $17$, then it iteratively merges these pairs to form final groups in lines $20$ to $52$.

\begin{algorithm}
	\caption{Pseudo-code of the Grouping Algorithm.}
	\label{alg:grouping}
	\KwIn{$TubeList; TH_D; TH_C$} 
	\KwOut{$GroupList$}
	$TubePair\gets Null$\;
	$N \gets Length(TubeList)$\;
	\For{$i \gets 1$ \KwTo $N$}
	{
		\For{$j \gets 1$ \KwTo $N$}
		{
			\If{$i\not=j$}
			{
				$T_1 \gets TubeList[i]; T_2 \gets TubeList[j]$\;
				$CF \gets CommonFrames(T_1, T_2)$\;
				$D \gets AverageDistance(T_1, T_2, CF)$\;
				$C \gets TotalCollision(T_1, T_2, CF)$\;
				$W \gets ConcurrencyWeight(T_1, T_2, CF)$\;
				$DW \gets D \times W$\;
				\If{$DW < TH_D$ \bfseries{or} $C > TH_C$}
				{
					$TubePair.append((T_1, T_2))$\;
				}	
			}				
		}
	}
	$GroupList\gets Null$\;
	\For{$i \gets 1$ \KwTo $N$}
	{
		$T_1 = TubePair[i][0]; T_2 = TubePair[i][1]$\;	
		$SkipGrouping = False$\;
		$M \gets Length(GroupList)$\;
		\For{$j \gets 1$ \KwTo $M$}
		{
			$G = GroupList[j]$\;
			\If{$(T1$ \bfseries{in} $G)$ \bfseries{or} $(T2$ in $G)$}
			{
				$SkipGrouping = True$\;
			}				
		}
		\If{$SkipGrouping$}
		{
			\bfseries{Continue}\;
		}
		$G = TubePair[i]$\;
		\For{$k \gets 1$ \KwTo $N$}
		{
			$StopExpansion = True$\;
			\For{$l \gets i+1$ \KwTo $N$}
			{
				$u = TubePair[l][0]; v = TubePair[l][1]$\;
				\If{$(u$ \bfseries{in} $G)$ \bfseries{and} $(v$ \bfseries{not in} $G)$}
				{
					$G.append(v)$\;
					$StopExpansion = False$\;				
				}
				\If{$(v$ \bfseries{in} $G)$ \bfseries{and} $(u$ \bfseries{not in} $G)$}
				{
					$G.append(u)$\;
					$StopExpansion = False$\;				
				}
			}
			\If{$StopExpansion$}
			{
				\bfseries{Break}\;
			}					
	  	}  
		$GroupList.append(G)$\;
	}
	$Sort\; GroupList\; by\; groups\; start\; in\; source\; video$\;
	\Return{$GroupList$}
\end{algorithm}

\subsubsection{Group Rearrangement}
Algorithm \ref{alg:Rearrangement} outlines the proposed group rearrangement method, which is a loop-based greedy algorithm. In each iteration, K groups are selected based on their start times in the source video, and their start times in the synopsis video are determined accordingly. This loop continues until all groups are processed.

To determine the start time of each group, an identical initial start time, named StartFrame, is assigned to each group. StartFrame is set to zero for the first iteration. For groups with multiple tubes, the tube with the earliest start time in the source video takes the StartFrame value, while other tubes within the same group take StartFrame plus their start time difference relative to the earliest tube.

In the subsequent step, the pairwise collision cost between each new group and all previously arranged groups is computed. If the collision cost between two groups exceeds a predefined threshold, the start time of the new group will be shifted forward by $3$ frames. This process of computing collision cost and shifting start times forward continues until the collision cost becomes less than the predefined threshold. It is worth noting that the collision cost of the new group is compared with previously arranged groups based on the order of their determined start times because the algorithm only shifts the new group forward. Optionally, multiple collision thresholds with varying shift amounts can be considered to increase the convergence rate.

Equation \ref{Groups collison} presents the deployed function used to calculate the collision cost between two groups:

\begin{equation}
\label{Groups collison}
C_G(G_i,G_j) = \frac{\sum_{m \in G_i} \sum_{n \in G_j} C(m,n)}{max(G_i,G_j)},
\end{equation}

\noindent where $C(m,n)$ denotes the function presented in Equation \ref{Group occlusion} used to compute the collision between tubes, and $max(G_i,G_j)$ represents the maximum number of bounding boxes contained in either of the two groups. The reason for using $max(G_i,G_j)$ is that as a group contains more bounding boxes, it is more prone to colliding with other groups. So, it is necessary to normalize the collision cost by the size of the groups. The same collision threshold is applied when shifting groups.

The challenge faced here is that, during pairwise collision cost computation, a collision between two groups can result in excessive shifts of one group. While these shifts are essential in the initial frames to avoid collisions, they become inefficient in the final frames, especially when the length of the output video increases only to prevent the collision of two groups. To address this issue, a collision weight parameter with an initial value of one is defined. This parameter scales the collision cost before comparing it with the collision threshold. If shifting a group increases the output video length, the collision weight is then multiplied by a decay rate to reduce its impact. This approach prevents excessive shifts of one group that would otherwise unnecessarily increase the video length.

As the rearrangement of groups progresses, the initial frames in the output video become entirely filled. Consequently, to arrange a new group, checking collision costs in these full frames is redundant since the new group must be shifted regardless. To optimize computations, the rearrangement of a new group begins from the frame indicated by the StartFrame variable. Details of the algorithm used to calculate the StartFrame are provided in the appendix.

\begin{algorithm}[!t]
\caption{Pseudo-code of the Group Rearrangement Algorithm.}
\label{alg:Rearrangement}
\KwIn{$GroupList; K; DecayRate; TH_{Cost}$}
\KwOut{$RGL$}
$VideoLength \gets Length\ of\ the\ longest\ group$\;
$RGL \gets Null$\;
$Sg \gets 0$\;
$Lg \gets Length(GroupList)$\;
$StartFrame \gets 0$\;
\While{$Sg < Lg$}
	{\For{$i \gets Sg$ \KwTo $ Min(Sg + K,Lg) $}
        {$GroupList[i].start = StartFrame$\;
        $Weight = 1$\;
		{\For{$j \gets 0$ \KwTo $Length(RGL)$}
			{$Cost=Collision(GroupList[i], RGL[j])$\;
			\If{$ Cost \times Weight > TH_{Cost} $}
			{$Shift(GroupList[i])$\;}
			\If{$GroupList[i].end > VideoLength$}
			{$VideoLength=GroupList[i].end$\;
			$Weight = Weight \times DecayRate$\;}
			}
		$RGL.append(GroupList[i])$\;
		$Sort\ RGL\ by\ groups\ start\ in\ synopsis\ video$\;
		}
	}
    $Sg = Sg + K$\;
    $StartFrame=CalculateStart(RGL)$\;
}
\Return{$RGL$}
\end{algorithm}

\subsection{Visualization}

\subsubsection{Background Generation}
During the tube extraction stage, when an empty frame is encountered, it is saved as a sample for background generation. Additionally, the system periodically saves a frame as a background sample after removing object pixels identified by the deep object detector. In the visualization stage, the system applies a pixel-wise median filter to the background samples to generate the background for the synopsis video.
 
\subsubsection{Segmentation}
Background pixels often appear within the bounding box of objects. Moreover, since tubes belong to different frames, there are typically slight differences between the background pixels inside the bounding box of objects and the generated background. So, placing the entire image of tubes on the extracted background would likely reduce the video's visual quality. Object segmentation emerges as a crucial solution to this challenge, ensuring precise separation of foreground objects from the background.

Object segmentation algorithms usually rely on extracting image features to identify object pixels, which can result in high computational costs. However, in scenarios with a static background, simpler algorithms can be utilized to detect object pixels more efficiently, reducing computational expenses. This work utilizes motion and the absolute difference between the extracted background and the image of each object's bounding box to eliminate background pixels within the bounding box. The absolute difference effectively distinguishes foreground from background pixels, as the intensity of a pixel from the extracted background closely matches its corresponding background pixel within an object bounding box, unlike a foreground pixel. However, some foreground pixels may be mistakenly detected as background due to similar colors to the background. Consequently, this paper also incorporates motion detection by calculating the absolute difference between two consecutive bounding boxes of a tube to determine object borders. 

For the generation of both masks, the absolute difference and motion are computed for RGB images, followed by averaging each channel to generate a one-channel mask. As each mask detects a portion of the foreground pixels, they are merged by summing their corresponding pixels. Variable thresholds are then employed to generate binary masks. Increasing the thresholds eliminates more background pixels, but it may also exclude some foreground pixels. To determine the threshold value, a high initial threshold is used, followed by an assessment of the ratio of pixels identified as foreground. A low ratio suggests that some foreground pixels have been overlooked, as the mask corresponds to the bounding box of an object. In response, the threshold is iteratively decreased until the ratio reaches an appropriate value.

\begin{figure}[!t]
	\centering
	\begin{tabular}{ccccc}
		\includegraphics[width=0.06\textwidth]{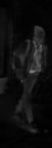} & \includegraphics[width=0.06\textwidth]{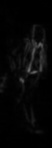} &
		\includegraphics[width=0.06\textwidth]{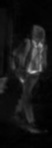} & \includegraphics[width=0.06\textwidth]{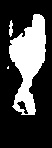} &
		\includegraphics[width=0.06\textwidth]{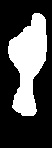}\\
		(a) & (b) & (c) & (d) & (e) \\
	\end{tabular}
	\caption{Different steps in segmentation mask generation: (a) Mask 1: Absolute difference between the extracted background and the object's bounding box image, (b) Mask 2: Motion, (c) Sum of the two masks, (d) Binary mask obtained using determined threshold, and (e) Final mask.}
	\label{Mask:fig}
\end{figure}

Subsequently, morphological operations are applied to reduce noise in the mask. Finally, utilizing the findContours function in OpenCV, the object's contour, which is the largest available contour in the mask, is identified. The output of each segmentation step is illustrated in Figure \ref{Mask:fig}.

\subsubsection{Stitching}
\begin{figure}[!t]
	\centering
	\begin{tabular}{ccc}
		\includegraphics[width=0.06\textwidth]{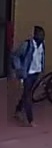} & \includegraphics[width=0.06\textwidth]{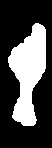} &
		\includegraphics[width=0.06\textwidth]{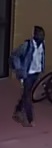}\\
		(a) & (b) & (c) \\
	\end{tabular}
	\caption{The impact of the segmentation mask on the quality of the generated synopsis video frame: (a) Placing the raw image of the object's bounding box on the background, (b) Segmentation mask, and (c) Placing segmented object pixels on the background.}
	\label{segmentation:fig}
\end{figure}

To generate each frame of the synopsis video, the extracted background is loaded initially. Then, for each object intended to be present in that particular frame, the segmentation mask is generated and applied, followed by the stitching of object pixels onto the background. An example demonstrating the impact of using the segmentation mask on video quality is depicted in Figure \ref{segmentation:fig}.

\section{Metrics}
Video synopsis performance can be affected by various factors. This research categorizes the comparison into four aspects:

\begin{enumerate}[leftmargin=*]
	\item \textbf{Compression and collision:} In video synopsis, the goal is to minimize video length for maximum compression while reducing collisions between tubes to avoid information loss. There is a trade-off between collision and compression, so both metrics should be considered when comparing synopsis models. This study proposes the frame condensation ratio (FR) and collision area (CA) parameters to measure compression and collision, respectively. FR is the ratio of synopsis video frames to source video frames. An FR of 1 indicates no compression, while a smaller FR indicates higher compression. CA refers to the total number of pixels in collisions between all pairs of objects. CA is determined by summing all the collisions of the bounding boxes in all frames of the synopsis video. An increase in CA indicates a greater loss of information.
	\item \textbf{Chronological order:} Maintaining the order of events in the video is crucial. In synopsis models, tubes should be arranged to preserve the video's event sequence. This work proposes the chronological disorder ratio (CDR) to evaluate chronological order preservation. CDR is the ratio of tube pairs in reverse order to the total number of tube pairs.		
	\item \textbf{Visual quality:} The visual quality of the synopsis video is influenced by factors such as the quality of the extracted background, object segmentation, and tube rearrangement. However, comparing the visual quality of different video synopsis models remains challenging, and this issue is a topic for future research.	
	\item \textbf{Computational cost:} Algorithm runtime and resource usage are critical considerations in video synopsis, particularly because it involves processing a continuous stream of source video in real-time.
\end{enumerate}

\section{Dataset}
This research introduces a video synopsis dataset, called SynoClip, consisting of uncrowded, lengthy videos captured from outdoor mounted cameras. The dataset comprises six videos, five of which were collected from YouTube \cite{Y1, Y2, Y3, Y4, Y5}  and one from the MEVA dataset \cite{Y6}. For each video, tubes were manually annotated to facilitate direct comparison of various models' performance. Video lengths range from $8$ to $45$ minutes. Table \ref{Dataset Table} details the specifications of each video.

\begin{table*}[!t]
	\def\arraystretch{1.5}
	\caption{Proposed dataset (SynoClip) specifications}
	\begin{center}
		\begin{tabular}{|c|c|c|c|c|c|c|c|c|} \hline
			Sequence & Resolution  & Length & Tube Count & Box Count	& Frame Count & Density & Coverage & Minimum FR \\ \hline
			Video1 & 1280$\times$720  & 08':11" & 141 & 78173  & 14743 & 2.90$\%$ & 0.58 & 0.196 \\ \hline
			Video2 & 768$\times$640   & 11':15" & 80  & 50704  & 20250 & 2.34$\%$ & 0.76 & 0.112 \\ \hline
			Video3 & 1280$\times$720  & 30':04" & 111 & 172900 & 54143 & 0.93$\%$ & 0.41 & 0.056 \\ \hline
			Video4 & 1800$\times$880  & 21':11" & 85  & 299571 & 38130 & 3.59$\%$ & 0.93 & 0.258 \\ \hline
			Video5 & 1650$\times$1030 & 11':21" & 201 & 141803 & 20423 & 3.57$\%$ & 0.62 & 0.288 \\ \hline
			Video6 & 1536$\times$1072 & 45':00" & 70  & 37638  & 81007 & 0.15$\%$ & 0.29 & 0.044 \\ \hline
		\end{tabular}
		\label{Dataset Table}
	\end{center}
\end{table*}

\begin{figure*}[!t]
	\centering
	\begin{tabular}{cc}
		\includegraphics[width=0.4\textwidth]{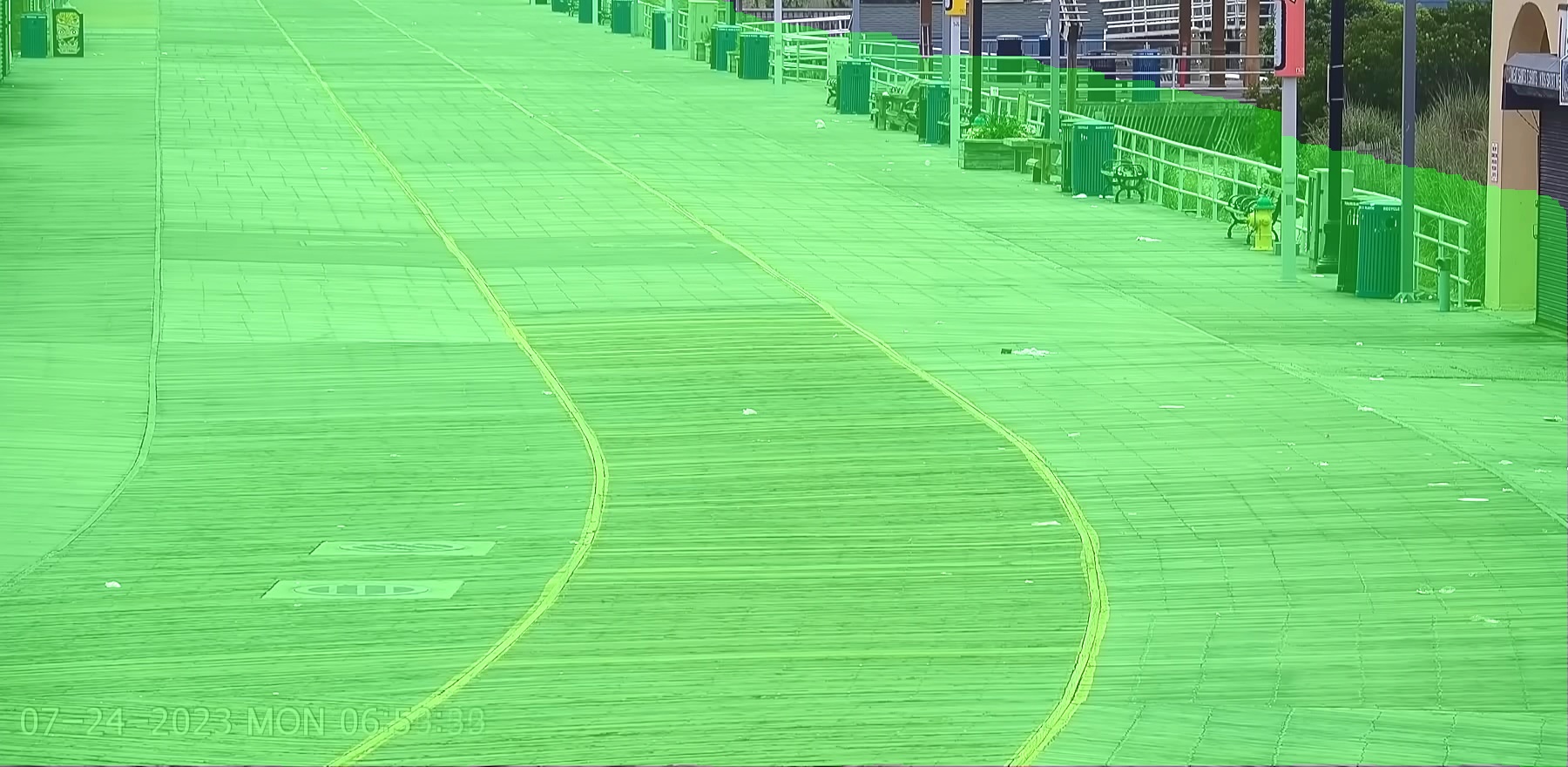} &
		 \includegraphics[width=0.3\textwidth]{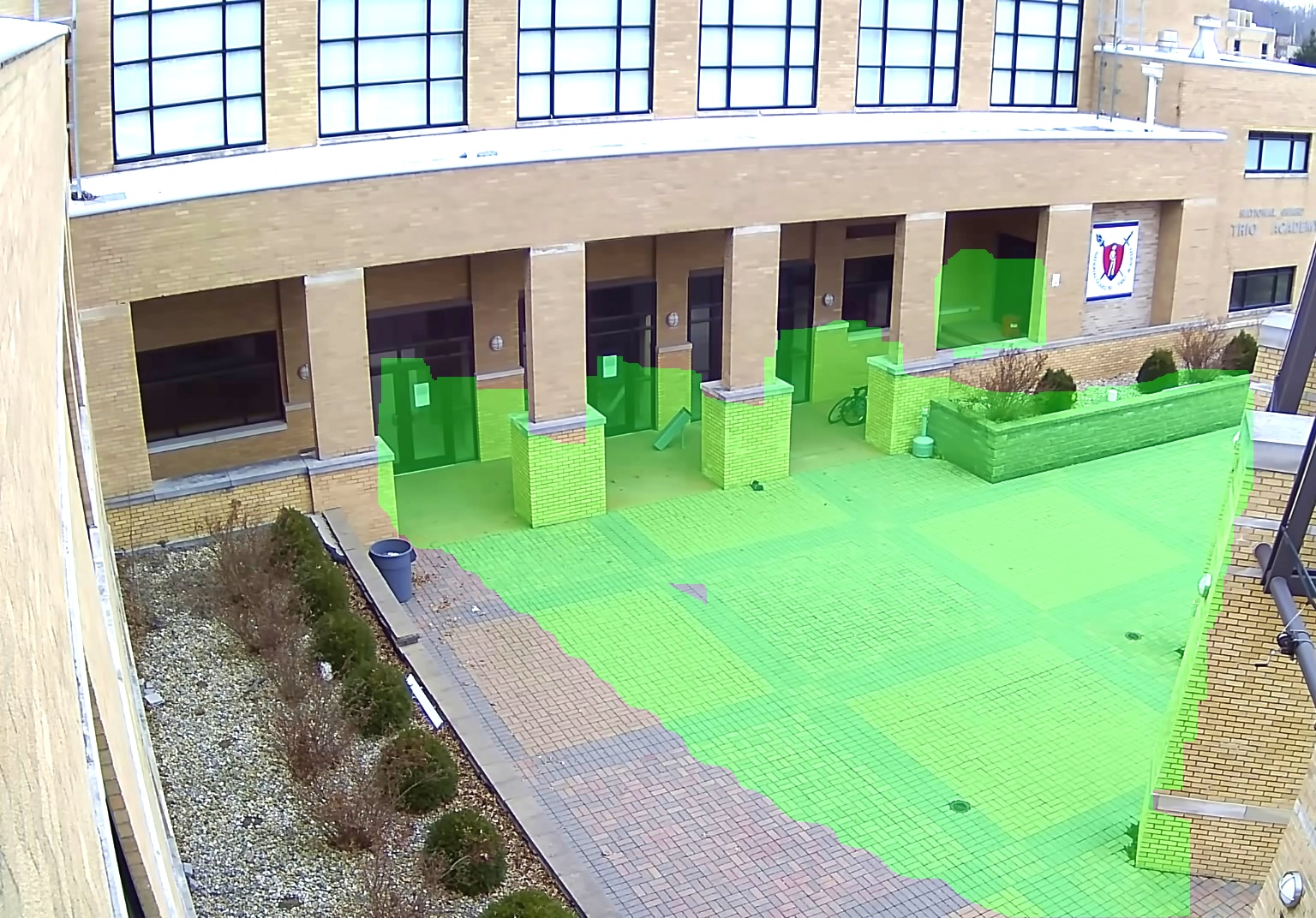} \\
		(a) & (b) \\
	\end{tabular}
	\caption{Coverage for two SynoClip sequences. Green areas represent pixels that contained objects at least once, indicating potential regions for object placement: (a) Video4: high coverage, with nearly the entire background occupied by objects over time. (b) Video6: low coverage, with a significant portion of the background taken up by buildings, limiting the available area for objects.}
	\label{Coverage:fig}
\end{figure*}

The first columns of Table \ref{Dataset Table} pertain to video specifications, while the last three are as follows:

\begin{itemize}[leftmargin=*]
	\item \textbf{Density:} Density indicates the crowdedness of the video, represented as the percentage of object box pixels in the video.
	\item \textbf{Coverage:} In the video synopsis task, the goal is to populate all frames with objects. However, there are often segments in the source video where no objects are present. Consequently, these areas cannot be filled with objects in the synopsis video. The Coverage column in Table \ref{Dataset Table} indicates the percentage of pixels where an object was present at least once during the video. Figure \ref{Coverage:fig} illustrates the coverage for two dataset sequences.
	\item \textbf{Minimum FR:} The length of the longest tube in each video determines the minimum achievable length of the synopsis video. The Minimum FR column in Table \ref{Dataset Table} shows the minimum frame condensation ratio (FR) achievable based on the length of the longest tube in the source video.
\end{itemize}

In what follows, the video specifications and synopsis generation challenges for each sequence in the SynoClip dataset are outlined:

\begin{enumerate}[leftmargin=*]
\item Video1: This video portrays a street during the day. There is no car movement observed in the footage. Two main pedestrian routes are visible, but their traffic varies significantly. This unbalanced load of tubes poses a challenge for the synopsis algorithm, potentially resulting in the final part of the synopsis video predominantly featuring tubes from one route. The density and coverage of this video are medium, with fewer lengthy tubes present. Moreover, it features a higher number of related tubes compared to other videos in the dataset. Additionally, object shadows present in this video pose challenges for background extraction and object segmentation tasks.

\item Video2: This video portrays a sidewalk during the day, where only one path is available for pedestrians. Shadows are not visible, and there are no obstacles causing object occlusions. The density of this video is low, but its coverage is high. Overall, this video poses fewer challenges compared to others in the dataset. The main challenge in this video arises from reflections of people in the windows of a building, which may lead to false object detections.

\item Video3: This video portrays a sidewalk during the night. It features medium density and coverage. The presence of numerous light sources complicates object detection and segmentation tasks.

\item Video4: This video portrays a sidewalk during the day and presents the highest coverage and density among the dataset, making it the most challenging. The average length of the tubes in this video is notably long. Although there are no physical obstacles, occlusions between tubes are frequent. A significant portion of the tubes involves people riding bicycles, leading to varying object speeds. Moreover, many tubes follow non-linear trajectories, which complicates both object tracking and the algorithm's task of determining optimal tube placement. Additionally, the presence of numerous birds in the video further complicates background generation tasks.

\item Video5: This video portrays a sidewalk during the day, featuring two main pedestrian routes. The density and coverage are relatively high. Two streetlights and a few moving cars cause occlusions, making object tracking challenging. Additionally, some people are riding bicycles, and the level of crowdedness changes over time. The presence of three lengthy tubes further adds to the challenges of this video.

\item Video6: This video portrays the front area of a building during the day. It has the lowest density and coverage in the dataset, with many frames not containing any objects. The length of this video is 45 minutes, making it the longest video in the dataset. The building's columns cause occlusions, posing challenges for the object tracking task. Additionally, sunlight varies over time, causing background changes. The video also contains two lengthy tubes.
\end{enumerate}

\section{Experiments}
To evaluate the performance of the proposed method, three experiments were conducted. The first experiment investigated the impact of the empty-frame object detector. In the second experiment, the tube rearrangement performance was evaluated using tube annotations. The results of this experiment can be directly compared with other studies that use the same annotation. The third experiment utilized the complete model to measure the speed of each component. All experiments were conducted using an Intel i7-6700K CPU and RTX 3080 Ti GPU, running on Python 3.10.12.

\subsection{Empty-frame Object Detector Impact}
The impact of the empty-frame object detector was evaluated using the Video3 and Video6 sequences, as the other videos contain objects in almost all frames. It should be noted that if all video frames contain objects, the empty-frame object detector will not be enabled and thus will not add computational overhead to the model.

To demonstrate the effectiveness of the empty-frame object detector, two experiments were conducted using the deep object detector: one including and one excluding the empty-frame object detector. The results of these experiments are summarized in Table \ref{Empty-frame Detector Table}.

\begin{table*}[!t]
	\begin{center}
		\caption{The impact of the empty-frame object detector on object detection performance}
		\begin{tabular}{cccccc}			
			\toprule
			Sequence & Empty-frames count & YOLO-only Runtime & {\bf Proposed Hybrid Method Runtime} & MOR & Empty-frame Object Detector Speed \\
			\midrule
			Video3 & 11652 & 533s & {\bf 433s} & 0.29$\%$ & 3025 FPS \\
			Video6 & 59516 & 963s & {\bf 327s} & 0.71$\%$ & 1947 FPS\\
			\bottomrule
		\end{tabular}
		\label{Empty-frame Detector Table}
	\end{center}
\end{table*}

To quantify the accuracy of the empty-frame object detector, Equation \ref{MOR} defines the Missed Object Rate (MOR). The MOR represents the percentage of objects that are missed because a frame is incorrectly identified as empty by the empty-frame object detector.

\begin{equation}
	\label{MOR}
	MOR = \frac{Number\; of\; missed\; objects}{Total\; number\; of\; objects}.
\end{equation}

As indicated in Table \ref{Empty-frame Detector Table}, less than $1\%$ of objects were lost in both videos, usually occurring when a person enters the scene and is only partially visible. Additionally, a reduction in object detection runtime of $23\%$ for Video3 and $66\%$ for Video6 was observed when using the empty-frame object detector. The last column of Table \ref{Empty-frame Detector Table} details the processing speed of the empty-frame object detector, which varies depending on the image resolution. Specifically, for Video3 with smaller dimensions, the detector achieved a speed of $3025$ frames per second, while for Video6, the speed was $1975$ frames per second.

\subsection{Tube Rearrangement Performance}
This experiment explores the performance of the tube rearrangement algorithm across various compression levels. This study aims to aggregate results from different videos to assess the overall model performance across the entire dataset. To achieve this, a spectrum of collision levels proportional to the total number of tube pixels in each video was specified. Specifically, CA levels were set at $2\%$, $4\%$, $6\%$, $8\%$, and $10\%$ of the total tube pixels. Given that the proposed model employs a pairwise collision threshold and CA cannot be predetermined, multiple collision thresholds were tested to approximate the target CA during model runs.

Table \ref{Compression Table} presents the findings of this experiment, displaying the FR and CA values for each video at various collision levels. The reported CA values reflect the model's output, which may slightly deviate from the target CA.
 
\begin{table*}[!t]
	\def\arraystretch{1.5}
	\caption{Compression and collision area at various collision levels}
	\begin{center}
		\begin{tabular}{|c| *{12}{c|}} \hline
			Collision
			& \multicolumn{2}{c|}{Video1}
			& \multicolumn{2}{c|}{Video2}
			& \multicolumn{2}{c|}{Video3}
			& \multicolumn{2}{c|}{Video4}
			& \multicolumn{2}{c|}{Video5}
			& \multicolumn{2}{c|}{Video6}\\ \cline{2-13}
			level & CA($\times 10^7$) & FR & CA($\times 10^7$) & FR & CA($\times 10^7$)
			& FR & CA($\times 10^7$) & FR & CA($\times 10^7$) & FR & CA($\times 10^7$)& FR\\
			\hline
			2$\%$  & 0.737 & 0.566 & 0.513 & 0.404 & 0.997 & 0.388
			& 4.210  & 0.551 & 2.457  & 0.680 & 0.414 & 0.065\\ 
			\hline
			4$\%$  & 1.663 & 0.419 & 0.993 & 0.368 & 1.878 & 0.306 
			& 8.329  & 0.410 & 4.271  & 0.550 & 0.752 & 0.054\\ 
			\hline
			6$\%$  & 2.373 &0.372 & 1.422 & 0.273 & 2.963 & 0.242 
			& 13.12 & 0.345 & 7.816  & 0.365 & 1.156 & 0.047\\
			\hline
			8$\%$  & 3.096 &0.312 & 1.877 & 0.225 & 3.766 & 0.201 
			& 16.95 & 0.321 & 9.889 & 0.335 & 1.612 & 0.046 \\ 
			\hline
			10$\%$ & 3.929 &0.278 & 2.390 & 0.188 & 4.599 & 0.178 
			& 18.06 & 0.298 & 13.13 & 0.322 & 1.997 & 0.045 \\ 
			\hline
		\end{tabular}
		\label{Compression Table}
	\end{center}
\end{table*}

To compare results across different videos by averaging FR values at each collision level, the FR values for each video must be unbiased. If FR values are biased, the model's performance in videos with higher FR values would be disproportionately emphasized, while its performance in videos with lower FR values would be underrepresented. Table \ref{Compression Table} highlights a significant bias in the FR values for each sequence, primarily due to differences in density and coverage across sequences. Lower coverage limits the ability to place tubes within a frame, while higher density requires more tube pixels to be placed in the synopsis video, both factors causing the FR value to increase. Therefore, to ensure a fair comparison, a standard measure must be established to normalize the FR value linearly with the coverage and density of each video. Equation \ref{normalize FR} introduces the normalized frame condensation ratio (NFR), a fair measure to compare the compression capabilities of different video synopsis models:

\begin{equation}
	\label{normalize FR}
	NFR = \frac{FR \times Coverage}{100\times Density}.
\end{equation}

\begin{figure*}[!t]
	\centering
	\begin{tabular}{cc}
		\includegraphics[width=0.42\textwidth]{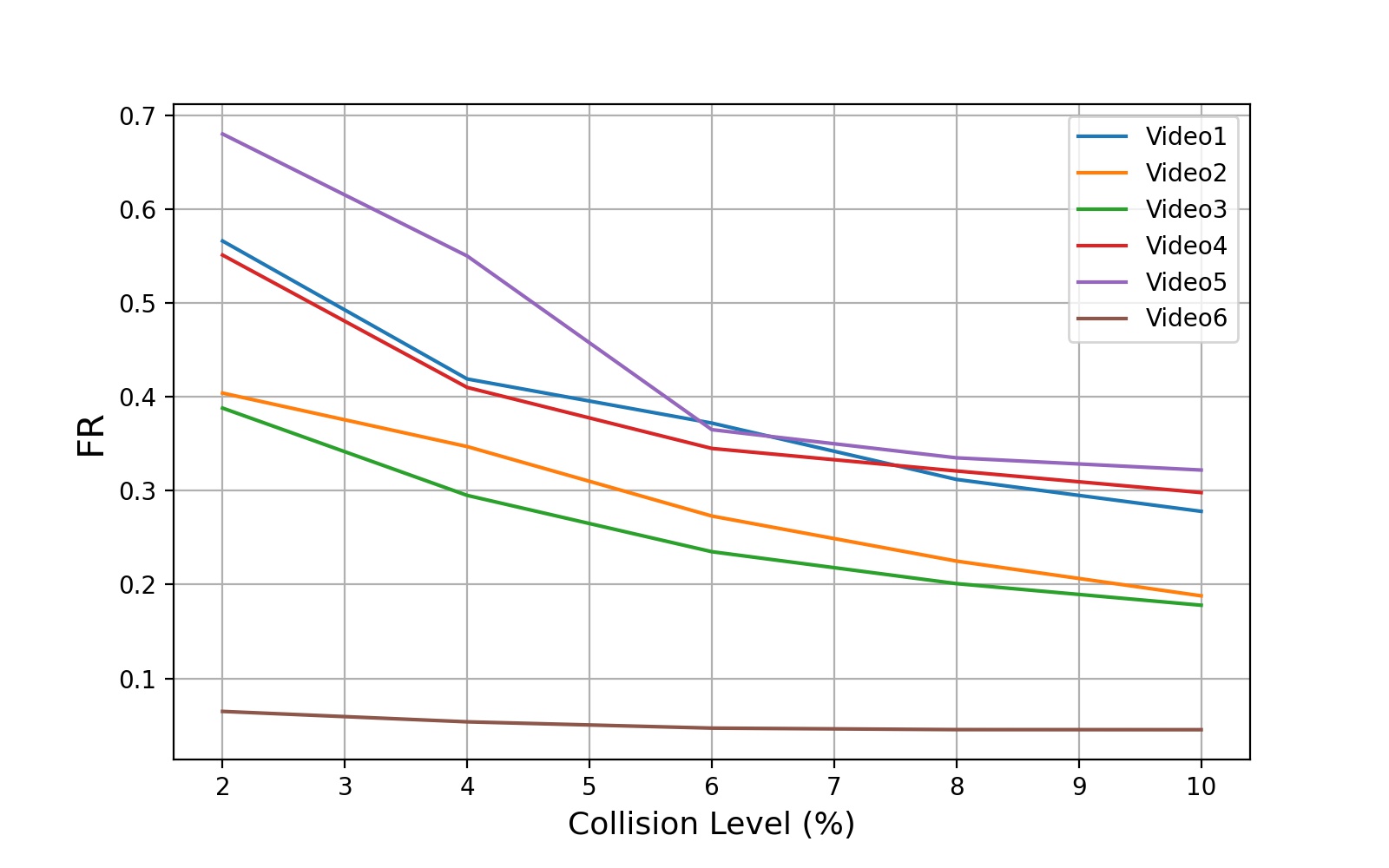} & 
		\includegraphics[width=0.42\textwidth]{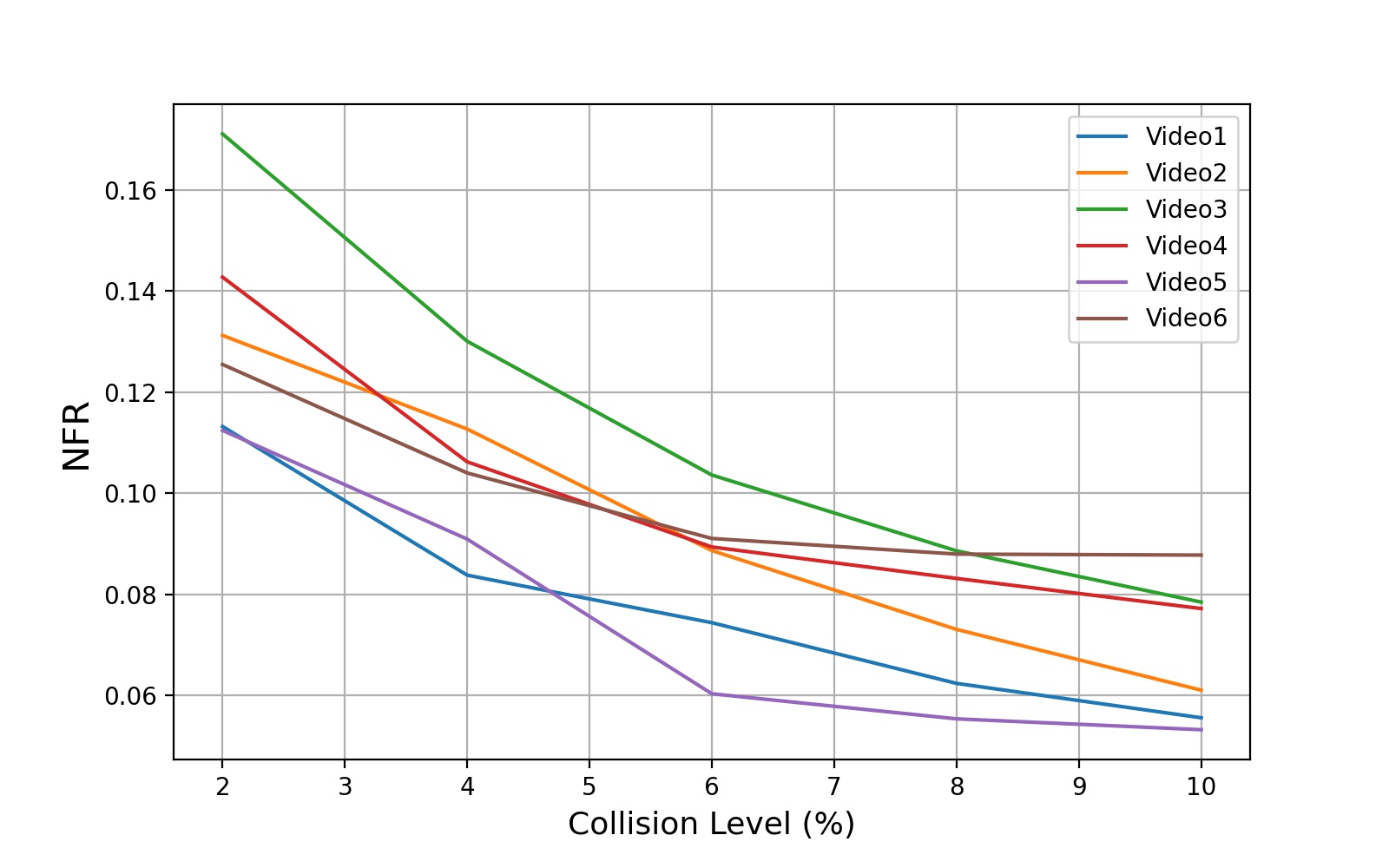} \\
		(a) & (b) \\
	\end{tabular}
	\caption{FR and NFR versus Collision Level for SynoClip Videos: (a) FR vs. Collision Level, (b) NFR vs. Collision Level}
	\label{NFR Fig}
\end{figure*}

In Equation \ref{normalize FR}, the FR value is determined after evaluating a synopsis model with a specific Collision level. Additionally, the Coverage and Density values for each video are available in Table \ref{Dataset Table}. Figure \ref{NFR Fig} displays the FR and NFR values at different Collision levels. The vertical axis of the curves shows that, at each Collision level, the variance of NFR across dataset videos is much lower compared to the variance of FR. Taking the average NFR at each collision level across all dataset videos will determine the overall compression rate of a synopsis model.

To provide a comprehensive evaluation of the tube rearrangement algorithm, it is recommended to measure compression, speed, and the capability of preserving the chronological order of events. Table \ref{Tube Table} shows the experiment results for the FGS. This table reports NFR to evaluate compression and CDR to evaluate chronological order preservation. It also reports speed in terms of the number of tubes processed per second. The numbers in Table \ref{Tube Table} represent the average results obtained from all dataset videos.

\begin{table}[!t]
	\begin{center}
		\def\arraystretch{1.5}
		\caption{Assessment of compression, chronological order preservation, and Speed at each collision level}
		\label{Tube Table}
		\begin{tabular}{c|ccc|ccc}
			\toprule
			\multirow{3}{*}{Collision level} &  \multicolumn{3}{c|}{\textbf{This work (FGS)}} & \multicolumn{3}{c}{\textbf{VSCS\cite{G1}}} \\
			  & NFR & CDR & Speed & NFR & CDR & Speed \\
			  &   &   & (TPS) &   &   & (TPS) \\
			\midrule
			2$\%$   & 0.133 & 0.174 & 6.761  & 0.169 & 0.457 & 0.608 \\
			4$\%$   & 0.105 & 0.172 & 7.752  & 0.154 & 0.449 & 0.773 \\
			6$\%$   & 0.085 & 0.175 & 10.08  & 0.144 & 0.434 & 0.964 \\
			8$\%$   & 0.075 & 0.186 & 12.40  & 0.137 & 0.426 & 1.069 \\
			10$\%$ & 0.069 & 0.181 & 13.93  & 0.128 & 0.425 & 1.174 \\
			\bottomrule
		\end{tabular}
	\end{center}
\end{table}

Most relevant studies have not shared their source code, which prevents direct comparisons. Therefore, this paper includes experiment results only for the VSCS model \cite{G1}. Table \ref{Tube Table} also shows the evaluation results for the VSCS model \cite{G1}. The proposed model, FGS, is inspired by the VSCS model, and there are similarities in the structure of the two methods. However, the FGS introduces four key improvements:

\begin{enumerate}[leftmargin=*]
	\item \textbf{Collision weight:} The FGS introduces a collision weight parameter that decreases as a group begins to extend the video's length.	
	\item \textbf{Pairwise group comparison:} In the VSCS model, new group allocation involves aggregating collision costs with all existing groups. If the total collision cost surpasses the threshold, the new group shifts. In contrast, the FGS computes collision costs pairwise with all preceding groups.	
	\item \textbf{Initial sorting:} The VSCS model initially sorts groups by their lengths in descending order to achieve better compression. However, with the introduction of a collision weight, good compression is achieved irrespective of the initial order. Consequently, the FGS sorts groups based on their start time in the source video to preserve the chronological order of events.	
	\item \textbf{Assignment of a new initial start time every K groups:} The VSCS model sets the initial time of all groups to zero. In contrast, the FGS calculates and assigns a new initial time after processing every K groups.
\end{enumerate}

These modifications enhance the FGS's performance across all three aspects of compression, chronological order preservation, and speed, as explained in what follows.

\begin{itemize}[leftmargin=*]
	\item \textbf{Compression:} The FGS model outperforms the VSCS model by incorporating the collision weight parameter. In the VSCS model, the lack of this parameter leads to excessive shifts in certain groups, resulting in unnecessary increases in video length. This distinction becomes more evident at higher compression rates, where the VSCS model tends to excessively fill the initial part of the output video while leaving significant unoccupied space at the end.
	\item \textbf{Chronological order preservation:} None of the models directly integrate the chronological order of tubes into the cost function. However, the FGS outperforms VSCS due to its method of processing groups. The VSCS model initially sorts groups by their length to achieve better compression, whereas the FGS sorts them based on their start times in the source video, thereby better preserving chronological order. While maintaining chronological order is independent of collision levels, the FGS shows approximately an $18\%$ alteration in the order of tubes, compared to a $44\%$ alteration when using the VSCS model.
	\item \textbf{Speed:} In the FGS, pairwise comparisons between groups significantly improve speed. Moreover, assigning a new initial time after processing every K groups further enhances speed, especially for videos with more tubes. The total runtime of both models depends on the total number of shifts. Consequently, speed increases as the collision level increases. Overall, as demonstrated in Table \ref{Tube Table}, the FGS model is approximately 10 times faster than the VSCS model.	
\end{itemize}

\subsection{Speed Evaluation of Each Component}
This experiment executes the complete model and measures the speed of different components. The results are presented in Table \ref{Speed}. In this table, the Runtime row indicates the percentage of total runtime consumed by each item. The speed of the tube extraction module primarily depends on video resolution, as higher resolutions slow down object detection. Video3 and Video6, which contain empty frames, demonstrate faster tube extraction speeds due to the use of empty-frame object detectors. As a result, while both Video5 and Video6 share the same video resolution, Video6 achieves nearly $3$ times faster object detection speed. The tracking algorithm \cite{sfsort} employed in the FGS is highly efficient, achieving a speed of $4392$ frames per second for the most crowded video in the dataset. The speed of the segmentation algorithm varies with the number of objects and video resolution. The FGS achieves an average speed of $1294$ frames per second across all dataset videos.

\begin{table}[!t]
	\begin{center}
		\def\arraystretch{1}
		\caption{The speed of each component per video (frames per second)}
		\label{Speed}
		\begin{tabular}{ccccc}
			\toprule
			Sequence & Detection & Tracking & Rearrangement & Segmentation \\
			\midrule
			Video1 & 285  & 5228   & 500     & 460   \\
			Video2 & 351  & 7500   & 2481    & 1012  \\
			Video3 & 378  & 7973   & 700     & 879   \\
			Video4 & 250  & 8200   & 465     & 433   \\
			Video5 & 215  & 4392   & 252     & 234   \\
			Video6 & 614  & 26073  & 17803   & 4748   \\
			\midrule
			Average & {\bf349} & {\bf9894} & {\bf3700} & {\bf1294} \\
			Runtime(\%) & {\bf71.4$\%$} & {\bf2.5$\%$} & {\bf6.8$\%$} & {\bf19.3$\%$} \\
			\bottomrule
		\end{tabular}
	\end{center}
\end{table}

Previous studies typically only reported the speed of their tube rearrangement modules \cite{G1, Nie, Nie2, Li, Ruan, swarm}. Among prior works, only SSOcT \cite{Yang} provided speeds for each component of their model processing a single video, with foreground extraction and tracking speeds of $51$ and $807$, respectively. The video resolution for Video1 and Video3 in SynoClip is identical to the video used in the SSOcT model experiment. The FGS achieves superior foreground extraction and tracking speeds of $176$ and $5228$ for Video1, and $264$ and $7973$ for Video3. This speed improvement is primarily attributed to the utilization of detection-with-stride and the empty-frame object detector. While these techniques have minimal impact on object detection accuracy, they significantly enhance object detection speed, which is the most time-consuming component, consuming $71.4\%$ of the total runtime, as shown in Table \ref{Speed}.

The speed of tube rearrangement depends entirely on the length and number of tubes in a video, resulting in significant variability based on video density. As indicated in Table \ref{Speed}, the tube rearrangement speed ranges from $252$ frames per second in the most crowded video to $17803$ frames per second in the least crowded one. Due to this variability, direct comparisons with other methods using different videos are not feasible. However, as shown in the table, tube rearrangement accounts for only $6.8\%$ of the total runtime, demonstrating the computational efficiency of the FGS.

\section{Conclusion}
This study introduces the FGS, a video synopsis model designed for low computational cost. In the tube extraction stage, an empty-frame object detector enhances model speed significantly in videos containing frames devoid of objects. The tube rearrangement stage employs a fast greedy algorithm that achieves high compression while preserving the relationship between tubes using a grouping algorithm. For visualization, a segmentation algorithm specifically designed for object bounding boxes removes background pixels within these boxes, thereby enhancing the quality of the synopsis video. Thanks to its efficiency, the FGS model achieves an average speed of $1294$ frames per second across all dataset videos. This research also introduces SynoClip, a standard video synopsis dataset comprising videos with essential attributes for evaluating synopsis models: they are lengthy, and captured from mounted cameras. Moreover, the dataset includes tube annotations, enabling direct comparison of performance across different methods. Furthermore, this paper introduces NFR as a fair measure to calculate model compression across various collision levels. At a collision level of $10\%$, compared to a comparable method, the FGS model operates $11.85$ times faster and achieves a video compression ratio $1.85$ times higher. Future work includes enabling the tube rearrangement algorithm to process video streams online. This can be achieved by adjusting the algorithm that calculates the initial start frame in the group rearrangement module.

\appendix 
\section*{Calculate StartFrame} 
After rearranging every K groups, a chart is created showing the number of bounding boxes detected in each frame of the output video. Each entry in this chart is based on the previously rearranged groups, referred to as the rearranged group list (RGL) in Algorithm \ref{alg:Rearrangement}. The maximum and average values from this chart are used to set a threshold for identifying the earliest section of the output video with sufficient space for additional groups. Therefore, for smaller values of K, it is preferable to select more groups in the first iteration of rearranging groups to obtain an accurate threshold from the chart. To calculate the StartFrame, the algorithm skips the first $15\%$ of the initial frames due to the typically small number of bounding boxes at the beginning, as tubes often originate from a few specific locations, leaving initial frames less populated. After skipping these initial frames, the algorithm identifies the number of the first frame with fewer bounding boxes than the threshold. Then, a fixed number is subtracted from this identified frame number to determine the StartFrame. This subtraction allows the algorithm to effectively fill this section of the output video, which had a low number of bounding boxes.


\begin{thebibliography}{1}
\bibliographystyle{IEEEtran}

\bibitem{VS1}
W.~Zhu, J.~Lu, J.~Li, and J.~Zhou, ``Dsnet: A flexible detect-to-summarize
  network for video summarization,'' \emph{IEEE Transactions on Image
  Processing}, vol.~30, pp. 948--962, 2021.

\bibitem{VS2}
T.~Liu, Q.~Meng, J.-J. Huang, A.~Vlontzos, D.~Rueckert, and B.~Kainz, ``Video
  summarization through reinforcement learning with a 3d spatio-temporal
  u-net,'' \emph{IEEE Transactions on Image Processing}, vol.~31, pp.
  1573--1586, 2022.

\bibitem{VS3}
T.-C. Hsu, Y.-S. Liao, and C.-R. Huang, ``Video summarization with
  spatiotemporal vision transformer,'' \emph{IEEE Transactions on Image
  Processing}, vol.~32, pp. 3013--3026, 2023.

\bibitem{VS4}
S.~Xiao, Z.~Zhao, Z.~Zhang, Z.~Guan, and D.~Cai, ``Query-biased self-attentive
  network for query-focused video summarization,'' \emph{IEEE Transactions on
  Image Processing}, vol.~29, pp. 5889--5899, 2020.

\bibitem{VS5}
W.~Zhu, Y.~Han, J.~Lu, and J.~Zhou, ``Relational reasoning over
  spatial-temporal graphs for video summarization,'' \emph{IEEE Transactions on
  Image Processing}, vol.~31, pp. 3017--3031, 2022.

\bibitem{VS6}
S.~Huang, X.~Li, Z.~Zhang, F.~Wu, and J.~Han, ``User-ranking video
  summarization with multi-stage spatio–temporal representation,'' \emph{IEEE
  Transactions on Image Processing}, vol.~28, no.~6, pp. 2654--2664, 2019.

\bibitem{VS7}
M.~Ma, S.~Mei, S.~Wan, Z.~Wang, X.-S. Hua, and D.~D. Feng, ``Graph
  convolutional dictionary selection with \( l_2 \), \( p \) norm for video
  summarization,'' \emph{IEEE Transactions on Image Processing}, vol.~31, pp.
  1789--1804, 2022.

\bibitem{Ruan}
T.~Ruan, S.~Wei, J.~Li, and Y.~Zhao, ``Rearranging online tubes for streaming
  video synopsis: A dynamic graph coloring approach,'' \emph{IEEE Transactions
  on Image Processing}, vol.~28, no.~8, pp. 3873--3884, 2019.

\bibitem{Yang}
Y.~Yang, H.~Kim, H.~Choi, S.~Chae, and I.-J. Kim, ``Scene adaptive online
  surveillance video synopsis via dynamic tube rearrangement using octree,''
  \emph{IEEE Transactions on Image Processing}, vol.~30, pp. 8318--8331, 2021.

\bibitem{G1}
X.~Li, Z.~Wang, and X.~Lu, ``Video synopsis in complex situations,'' \emph{IEEE
  Transactions on Image Processing}, vol.~27, no.~8, pp. 3798--3812, 2018.

\bibitem{Zhu}
J.~Zhu, S.~Feng, D.~Yi, S.~Liao, Z.~Lei, and S.~Z. Li, ``High-performance video
  condensation system,'' \emph{IEEE Transactions on Circuits and Systems for
  Video Technology}, vol.~25, no.~7, pp. 1113--1124, 2015.

\bibitem{Fu}
W.~Fu, J.~Wang, L.~Gui, H.~Lu, and S.~Ma, ``Online video synopsis of structured
  motion,'' \emph{Neurocomputing}, vol. 135, pp. 155--162, 2014.

\bibitem{Pritch}
Y.~Pritch, A.~Rav-Acha, and S.~Peleg, ``Nonchronological video synopsis and
  indexing,'' \emph{IEEE Transactions on Pattern Analysis and Machine
  Intelligence}, vol.~30, no.~11, pp. 1971--1984, 2008.

\bibitem{Ghatak}
S.~Ghatak, S.~Rup, B.~Majhi, and M.~N.~S. Swamy, ``Hsajaya: An improved
  optimization scheme for consumer surveillance video synopsis generation,''
  \emph{IEEE Transactions on Consumer Electronics}, vol.~66, no.~2, pp.
  144--152, 2020.

\bibitem{swarm}
M.~M. Moussa and R.~Shoitan, ``Object-based video synopsis approach using
  particle swarm optimization,'' \emph{Signal, Image and Video Processing},
  vol.~15, no.~4, pp. 761--768, 2021.

\bibitem{Li}
X.~Li, Z.~Wang, and X.~Lu, ``Surveillance video synopsis via scaling down
  objects,'' \emph{IEEE Transactions on Image Processing}, vol.~25, no.~2, pp.
  740--755, 2016.

\bibitem{Nie}
Y.~Nie, Z.~Li, Z.~Zhang, Q.~Zhang, T.~Ma, and H.~Sun, ``Collision-free video
  synopsis incorporating object speed and size changes,'' \emph{IEEE
  Transactions on Image Processing}, vol.~29, pp. 1465--1478, 2020.

\bibitem{Nie2}
Y.~Nie, C.~Xiao, H.~Sun, and P.~Li, ``Compact video synopsis via global
  spatiotemporal optimization,'' \emph{IEEE Transactions on Visualization and
  Computer Graphics}, vol.~19, no.~10, pp. 1664--1676, 2013.

\bibitem{Chen}
S.~Chen, X.~Liu, Y.~Huang, C.~Zhou, and H.~Miao, ``Video synopsis based on
  attention mechanism and local transparent processing,'' \emph{IEEE Access},
  vol.~8, pp. 92\,603--92\,614, 2020.

\bibitem{Huang}
C.-R. Huang, P.-C.~J. Chung, D.-K. Yang, H.-C. Chen, and G.-J. Huang, ``Maximum
  a posteriori probability estimation for online surveillance video synopsis,''
  \emph{IEEE Transactions on Circuits and Systems for Video Technology},
  vol.~24, no.~8, pp. 1417--1429, 2014.

\bibitem{Jin}
J.~Jin, F.~Liu, Z.~Gan, and Z.~Cui, ``Online video synopsis method through
  simple tube projection strategy,'' in \emph{2016 8th International Conference
  on Wireless Communications \& Signal Processing (WCSP)}, 2016, pp. 1--5.

\bibitem{Feng}
S.~Feng, Z.~Lei, D.~Yi, and S.~Z. Li, ``Online content-aware video
  condensation,'' in \emph{2012 IEEE Conference on Computer Vision and Pattern
  Recognition}, 2012, pp. 2082--2087.

\bibitem{He}
Y.~He, Z.~Qu, C.~Gao, and N.~Sang, ``Fast online video synopsis based on
  potential collision graph,'' \emph{IEEE Signal Processing Letters}, vol.~24,
  no.~1, pp. 22--26, 2017.

\bibitem{Ra}
M.~Ra and W.-Y. Kim, ``Parallelized tube rearrangement algorithm for online
  video synopsis,'' \emph{IEEE Signal Processing Letters}, vol.~25, no.~8, pp.
  1186--1190, 2018.

\bibitem{PETS}
B.~Yang and R.~Nevatia, ``Multi-target tracking by online learning of
  non-linear motion patterns and robust appearance models,'' in \emph{2012 IEEE
  Conference on Computer Vision and Pattern Recognition}, 2012, pp. 1918--1925.

\bibitem{CAVIAR}
R.~R. Sillito and B.~Fisher, ``Semi-supervised learning for anomalous
  trajectory detection,'' in \emph{Proceedings British Machine Vision
  Conference BMVC2008}, 2008, pp. 1035--1044.

\bibitem{Hall}
T.~Mahalingam and M.~Subramoniam, ``Aco--mkfcm: an optimized object detection
  and tracking using dnn and gravitational search algorithm,'' \emph{Wireless
  Personal Communications}, vol. 110, no.~3, pp. 1567--1604, 2020.

\bibitem{Daytime}
B.~Kille, F.~Hopfgartner, T.~Brodt, and T.~Heintz, ``The plista dataset,'' in
  \emph{Proceedings of the 2013 international news recommender systems workshop
  and challenge}, 2013, pp. 16--23.

\bibitem{F}
T.~Wang, J.~Liang, X.~Wang, and S.~Wang, ``Background modeling using local
  binary patterns of motion vector,'' in \emph{2012 Visual Communications and
  Image Processing}, 2012, pp. 1--5.

\bibitem{KTH}
C.~Schuldt, I.~Laptev, and B.~Caputo, ``Recognizing human actions: a local svm
  approach,'' in \emph{Proceedings of the 17th International Conference on
  Pattern Recognition, 2004. ICPR 2004.}, vol.~3, 2004, pp. 32--36 Vol.3.

\bibitem{WEIZMAN}
M.~Blank, L.~Gorelick, E.~Shechtman, M.~Irani, and R.~Basri, ``Actions as
  space-time shapes,'' in \emph{Tenth IEEE International Conference on Computer
  Vision (ICCV'05) Volume 1}, vol.~2, 2005, pp. 1395--1402 Vol. 2.

\bibitem{VIRAT}
S.~Oh, A.~Hoogs, A.~Perera, N.~Cuntoor, C.-C. Chen, J.~T. Lee, S.~Mukherjee,
  J.~K. Aggarwal, H.~Lee, L.~Davis, E.~Swears, X.~Wang, Q.~Ji, K.~Reddy,
  M.~Shah, C.~Vondrick, H.~Pirsiavash, D.~Ramanan, J.~Yuen, A.~Torralba,
  B.~Song, A.~Fong, A.~Roy-Chowdhury, and M.~Desai, ``A large-scale benchmark
  dataset for event recognition in surveillance video,'' in \emph{CVPR 2011},
  2011, pp. 3153--3160.

\bibitem{Sherbrooke}
J.-P. Jodoin, G.-A. Bilodeau, and N.~Saunier, ``Urban tracker: Multiple object
  tracking in urban mixed traffic,'' in \emph{IEEE Winter Conference on
  Applications of Computer Vision}, 2014, pp. 885--892.

\bibitem{D1}
S.~Jagtap and N.~B. Chopade, ``A comprehensive investigation about video
  synopsis methodology and research challenges,'' \emph{Inventive Computation
  and Information Technologies: Proceedings of ICICIT 2020}, pp. 911--923,
  2021.

\bibitem{D2}
K.~B. Baskurt and R.~Samet, ``Video synopsis: A survey,'' \emph{Computer Vision
  and Image Understanding}, vol. 181, pp. 26--38, 2019.

\bibitem{D3}
N.~K. and A.~Narayanan, ``Video synopsis: State-of-the-art and research
  challenges,'' in \emph{2018 International Conference on Circuits and Systems
  in Digital Enterprise Technology (ICCSDET)}, 2018, pp. 1--10.

\bibitem{D4}
P.~Y. Ingle and Y.-G. Kim, ``Video synopsis algorithms and framework: A survey
  and comparative evaluation,'' \emph{Systems}, vol.~11, no.~2, p. 108, 2023.

\bibitem{yolo7}
C.-Y. Wang, A.~Bochkovskiy, and H.-Y.~M. Liao, ``Yolov7: Trainable
  bag-of-freebies sets new state-of-the-art for real-time object detectors,''
  in \emph{2023 IEEE/CVF Conference on Computer Vision and Pattern Recognition
  (CVPR)}, 2023, pp. 7464--7475.

\bibitem{yolo8}
[Online]. Available: {\url{https://github.com/ultralytics/ultralytics}}.

\bibitem{sfsort}
M.M. ~Morsali, Z.~Sharifi, F.~Fallah, S.~Hashembeiki, H.~Mohammadzade, and
  S.B. ~Shouraki, ``Sfsort: Scene features-based simple online real-time
  tracker,'' \emph{arXiv preprint arXiv:2404.07553}, 2024.

\bibitem{Y1}
Faisal Imtiaz. Pedestrian Walking ,Human Activity Recognition Video ,DataSet By UET Peshawar. (Nov 2, 2016). Accessed Jul. 16, 2023. [Online video]. Available: {\url{https://www.youtube.com/watch?v=2bKXv\_XviFc}}

\bibitem{Y2}
Tanukiya LIVE. [LIVE CAMERA] Tanukikoji Shopping Street in Sapporo, Hokkaido, Japan. (2021). Accessed: Jul. 15, 2023. [Streaming Video]. Available: {\url{https://www.youtube.com/watch?v=ybWlQJfQrwM}}

\bibitem{Y3}
Kranjska Gora. Kranjska Gora Town Center - Live WebCam. (2022). Accessed: Sep. 5, 2023. [Streaming Video]. Available: {\url{https://www.youtube.com/watch?v=W27kWZAwCH4}}

\bibitem{Y4}
APM Digital. AC Boardwalk Live. (2023). Accessed: Jul. 24, 2023. [Streaming Video]. Available: {\url{https://www.youtube.com/watch?v=RIA5ekgxetE}}

\bibitem{Y5}
Radical Video Jockey-RVJJP. [LIVE] Osaka Dotonbori Live Camera. (2022). Accessed: Jul. 25, 2023. [Streaming Video]. Available: {\url{https://www.youtube.com/watch?v=sbSKv5U0tAc}}

\bibitem{Y6}
D.~Davila, D.~Du, B.~Lewis, C.~Funk, J.~Van~Pelt, R.~Collins, K.~Corona,
  M.~Brown, S.~McCloskey, A.~Hoogs, and B.~Clipp, ``Mevid: Multi-view extended
  videos with identities for video person re-identification,'' in \emph{2023
  IEEE/CVF Winter Conference on Applications of Computer Vision (WACV)}, 2023,
  pp. 1634--1643.

\end{thebibliography}
\end{document}